\documentclass[10pt,twocolumn,letterpaper]{article}

\usepackage{cvpr}
\usepackage{times}
\usepackage{epsfig}
\usepackage{graphicx}
\usepackage{amsmath}
\usepackage{amssymb}
\usepackage{bbm}

\usepackage{subcaption}
\usepackage{caption}
\usepackage{color}
\usepackage{xcolor,colortbl}
\usepackage{capt-of}
\usepackage{wrapfig}
\usepackage{tabularx}
\usepackage{booktabs}

\usepackage{amsmath,amsfonts,amsthm,amssymb,bm}

\def\eqref#1{equation~\ref{#1}}

\def\1{\bm{1}}

\DeclareMathAlphabet{\mathsfit}{\encodingdefault}{\sfdefault}{m}{sl}
\SetMathAlphabet{\mathsfit}{bold}{\encodingdefault}{\sfdefault}{bx}{n}

\DeclareMathOperator*{\argmax}{arg\,max}

\newcolumntype{C}[1]{>{\centering\let\newline\\\arraybackslash\hspace{0pt}}p{#1}}

\newcommand{\todo}[1]{{\color{red}{\small\bf\sf [TODO:#1]}}}
\newcommand{\wk}[1]{{\color{magenta}{\small\bf\sf [wk:#1]}}}
\newcommand{\wkk}[1]{{\color{magenta}#1}}

\newcommand{\red}[1]{{\color{red}#1}}

\newcommand*{\minl}{\min\limits}
\newcommand*{\maxl}{\max\limits}
\newcommand{\Ex}{\mathbb{E}}
\newcommand{\KLD}{\text{D}_{\text{KL}}}
\newcommand{\JSD}{\text{D}_{\text{JS}}}
\newcommand{\regours}{\mathcal{R}_{\text{ID}}}
\newcommand{\regvae}{\mathcal{R}_{\text{VAE}}}
\newcommand{\reginfo}{\mathcal{R}_{\text{Info}}}
\newcommand{\ours}{ID-GAN}

\newcommand{\cutabstractdown}{\vspace*{-0.2in}}

\newcommand{\cutsectiondown}{\vspace*{-0.08in}}
\newcommand{\cutsubsectionup}{\vspace*{-0.05in}}
\newcommand{\cutsubsectiondown}{\vspace*{-0.07in}}
\newcommand{\cutparagraphup}{\vspace*{-0.08in}}

\definecolor{Gray}{gray}{0.85}
\newcolumntype{a}{>{\columncolor{Gray}}c}

\usepackage[pagebackref=true,breaklinks=true,letterpaper=true,colorlinks,bookmarks=false]{hyperref}

\cvprfinalcopy %
\def\cvprPaperID{1315} %

\ifcvprfinal\fi
\begin{document}

\title{High-Fidelity Synthesis with Disentangled Representation\vspace{-0.2cm}}

\author{
\begin{tabular}{cccc}
Wonkwang Lee\textsuperscript{1}& Donggyun Kim\textsuperscript{1}& Seunghoon Hong\textsuperscript{1}& Honglak Lee\textsuperscript{2,3}
\vspace{0.25cm}
\end{tabular}\\
\begin{tabular}{@{}C{4.5cm}@{}C{4.5cm}@{}C{4.5cm}}
\textsuperscript{1}KAIST & \textsuperscript{2}University of Michigan & \textsuperscript{3}Google Brain \\
\end{tabular}\\
\vspace{0.2cm}
\begin{tabular}{cc}
{\tt\small \textsuperscript{1}\{wonkwang.lee,kdgyun425,seunghoon.hong\}@kaist.ac.kr} & \textsuperscript{2,3}{\tt\small honglak@\{umich.edu,google.com\}}
\end{tabular}
\vspace{-0.2cm}
}

\twocolumn[{%
\renewcommand\twocolumn[1][]{#1}%
\vspace{-3.5em}
\maketitle
\vspace{-2.5em}
\begin{center}
  \centering
  \includegraphics[width=0.999\linewidth]{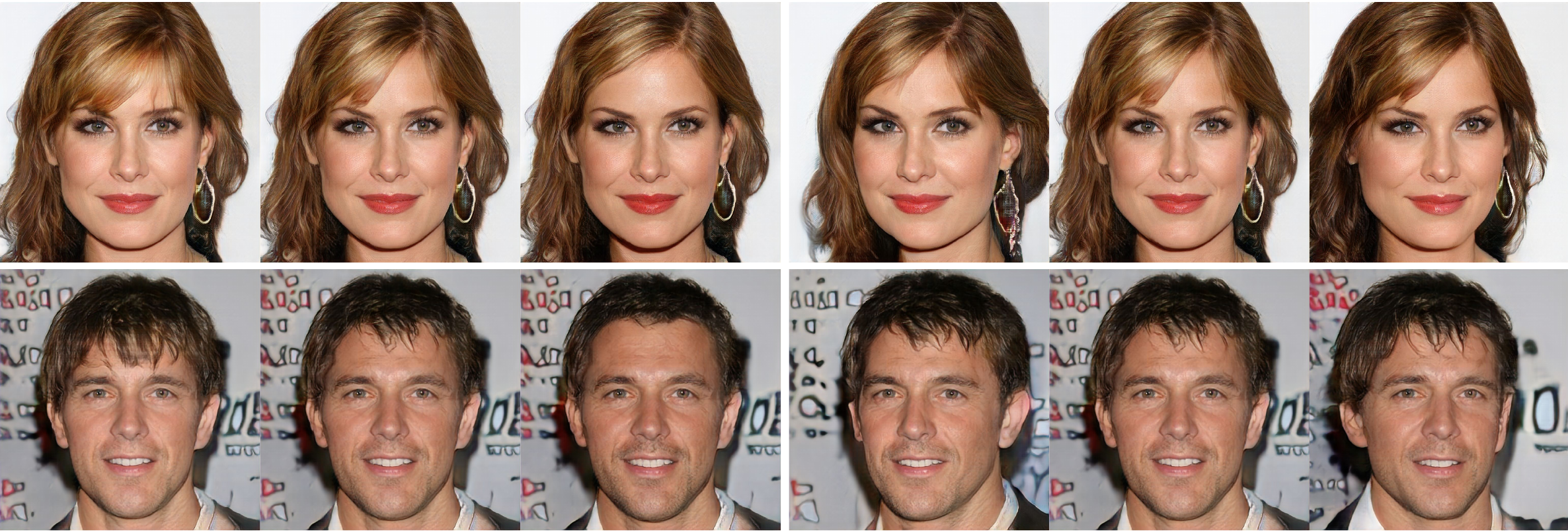}
  \vspace*{-20pt}
  \captionof{figure}{
  Generated images on the CelebA-HQ dataset~\cite{dsr/celeba}. The proposed framework allows synthesizing high-resolution images (1024$\times$1204 pixels) using the disentangled representation learned by VAEs.
}
  \label{fig:hq_synthesis}
\end{center}%
}]

\begin{abstract}
Learning disentangled representation of data without supervision is an important step towards improving the interpretability of generative models.
Despite recent advances in disentangled representation learning, existing approaches often suffer from the trade-off between representation learning and generation performance (\ie~improving generation quality sacrifices disentanglement performance). 
We propose an Information-Distillation Generative Adversarial Network (ID-GAN), a simple yet generic framework that easily incorporates the existing state-of-the-art models for both disentanglement learning and high-fidelity synthesis.
Our method learns disentangled representation using VAE-based models, and distills the learned representation with an additional nuisance variable to the separate GAN-based generator for high-fidelity synthesis. 
To ensure that both generative models are aligned to render the same generative factors, we further constrain the GAN generator to maximize the mutual information between the learned latent code and the output. 
Despite the simplicity, we show that the proposed method is highly effective, achieving comparable image generation quality to the state-of-the-art methods using the disentangled representation.
We also show that the proposed decomposition leads to an efficient and stable model design, and we demonstrate photo-realistic high-resolution image synthesis results (1024x1024 pixels) for the first time using the disentangled representations.
\end{abstract}
\cutabstractdown

\section{Introduction}
\label{sec:intro}
\vspace*{-0.1in}

Learning a compact and interpretable representation of data without supervision is important to improve our understanding of data and machine learning systems.
Recently, it is suggested that a \emph{disentangled representation}, which represents data using independent factors of variations in data can improve the interpretability and transferability of the representation~\cite{misc/bengio, dsr/info-dropout, misc/recentadv}. 
Among various use-cases of disentangled representation, we are particularly interested in its application to generative models, since it allows users to specify the desired properties in the output by controlling the generative factors encoded in each latent dimension.
There are increasing demands on such generative models in various domains, such as image manipulation~\cite{msic/MUNIT, misc/DRIT, misc/sup_att_fader}, drug discovery~\cite{misc/drugVAE}, ML fairness~\cite{misc/fairness2, misc/fairness}, \etc. 

Most prior works on unsupervised disentangled representation learning formulate the problem as constrained generative modeling task. 
Based on well-established frameworks, such as the Variational Autoencoder (VAE) or the Generative Adversarial Network (GAN), they introduce additional regularization to encourage the axes of the latent manifold to align with independent generative factors in the data.
Approaches based on VAE~\cite{dsr/betavae,dsr/betatcvae,dsr/factorvae,dsr/annealedvae} augment its objective function to favor a factorized latent representation by adding implicit~\cite{dsr/betavae,dsr/annealedvae} or explicit penalties~\cite{dsr/factorvae,dsr/betatcvae}.
On the other hand, approaches based on GAN~\cite{dsr/infogan} propose to regularize the generator such that it increases the mutual information between the input latent code and its output.

One major challenge in the existing approaches is the trade-off between learning disentangled representations and generating realistic data. 
VAE-based approaches are effective in learning useful disentangled representations in various tasks, but their generation quality is generally worse than the state-of-the-arts, which limits its applicability to the task of realistic synthesis.
On the other hand, GAN-based approaches can achieve the high-quality synthesis with a more expressive decoder and without explicit likelihood estimation~\cite{dsr/infogan}.
However, they tend to learn comparably more entangled representations than the VAE counterparts~\cite{dsr/betavae, dsr/factorvae, dsr/betatcvae, dsr/annealedvae} and are notoriously difficult to train, even with recent techniques to stabilize the training~\cite{dsr/factorvae, misc/infowgangp}. 

To circumvent this trade-off, we propose a simple and generic framework to combine the benefits of disentangled representation learning and high-fidelity synthesis.
Unlike the previous approaches that address both problems jointly by a single objective, we formulate two separate, but successive problems; we first learn a disentangled representation using VAE, and \emph{distill} the learned representation to GAN for high-fidelity synthesis. 
The distillation is performed from VAE to GAN by transferring the inference model, which provides a meaningful latent distribution, rather than a simple Gaussian prior and ensures that both models are aligned to render the same generative factors.
Such decomposition also naturally allows a layered approach to learn latent representation by first learning major disentangled factors by VAE, then learning missing (entangled) nuisance factors by GAN. 
We refer the proposed method as the Information Distillation Generative Adversarial Network (ID-GAN).

Despite the simplicity, the proposed ID-GAN is extremely effective in addressing the previous challenges, achieving high-fidelity synthesis using the learned disentangled representation (\eg~1024$\times$1024 image). 
We also show that such decomposition leads to a practically efficient model design, allowing the models to learn the disentangled representation from low-resolution images and transfer it to synthesize high-resolution images.  

The contributions of this paper are as follows:
\begin{itemize}
    \item We propose ID-GAN, a simple yet effective framework that combines the benefits of disentangled representation learning and high-fidelity synthesis.
    \item The decomposition of the two objectives enables plug-and-play-style adoption of state-of-the-art models for both tasks, 
    and efficient training by learning models for disentanglement and synthesis using low- and high-resolution images, respectively.
    \item Extensive experimental results show that the proposed method achieves state-of-the-art results in both disentangled representation learning and synthesis over a wide range of tasks from synthetic to complex datasets.
\end{itemize}

\section{Related Work}
\label{sec:related_work}
\cutsectiondown
\paragraph{Disentanglement learning.}
Unsupervised disentangled representation learning aims to discover a set of generative factors, whose element encodes unique and independent factors of variation in data.
To this end, most prior works based on VAE~\cite{dsr/betavae,dsr/factorvae,dsr/betatcvae} and GAN~\cite{dsr/infogan, dsr/ibgan,dsr/oogan, dsr/infogancr} focused on designing the loss function to encourage the factorization of the latent code.
Despite some encouraging results, however, these approaches have been mostly evaluated on simple and low-resolution images~\cite{dsr/dsprites, dsr/google}.
We believe that improving the generation quality of disentanglement learning is important, since it not only increases the practical impact in real-world applications, but also helps us to better assess the disentanglement quality on complex and natural images where the quantitative evaluation is difficult.
Although there are increasing recent efforts to improve the generation quality with disentanglement learning~\cite{dsr/ibgan,misc/hologan,dsr/oogan,dsr/infogancr}, they often come with the degraded disentanglement performance~\cite{dsr/infogan}, rely on a specific inductive bias (\eg~3D transformation~\cite{misc/hologan}), or are limited to low-resolution images~\cite{dsr/oogan, dsr/infogancr, dsr/ibgan}. 
On the contrary, our work aims to investigate a general framework to improve the generation quality without representation learning trade-off, while being general enough to incorporate various methods and inductive biases.
We emphasize that this contribution is complementary to the recent efforts for designing better inductive bias or supervision for disentanglement learning~\cite{dsr/sbd, dsr/semi1, dsr/semi2, dsr/semi3, dsr/semi4}. 
In fact, our framework is applicable to a wide variety of disentanglement learning methods and can incorporate them in a plug-and-play style as long as they have an inference model (\eg~nonlinear ICA~\cite{dsr/ivae}).

\cutparagraphup
\paragraph{Combined VAE/GAN models.}
There have been extensive attempts in literature toward building hybrid models of VAE and GAN~\cite{misc/vaegan,misc/cvaegan,misc/introvae,misc/bicyclegan,misc/ian}.
These approaches typically learn to represent and synthesize data by combining VAE and GAN objectives and optimizing them jointly in an end-to-end manner. 
Our method is an instantiation of this model family, but is differentiated from the prior work in that (1) the training of VAE and GAN is decomposed into two separate tasks and (2) the VAE is used to learn a specific conditioning variable (\ie~disentangled representation) to the generator while the previous methods assume the availability of an additional conditioning variable~\cite{misc/cvaegan} or use VAE to learn the entire (entangled) latent distribution~\cite{misc/vaegan, misc/introvae, misc/bicyclegan,misc/ian}. 
In addition, extending the previous VAE-GAN methods to incorporate disentanglement constraints is not straightforward, as the VAE and GAN objectives are tightly entangled in them. 
In the experiment, we demonstrate that applying existing hybrid models on our task typically suffers from the suboptimal trade-off between the generation quality and the disentanglement performance, and they perform much worse than our method.

\section{Background: Disentanglement Learning}
\label{sec:background}
\cutsectiondown
The objective of unsupervised disentanglement learning is to describe each data $x$ using a set of statistically independent generative factors $z$.
In this section, we briefly review prior works and discuss their advantages and limitations.

The state-of-the-art approaches in unsupervised disentanglement learning are largely based on the Variational Autoencoder (VAE).
They rewrite their original objective and derive regularizations that encourage the disentanglement of the latent variables.
For instance, $\beta$-VAE~\cite{dsr/betavae} proposes to optimize the following modified Evidence Lower-Bound~(ELBO) of the marginal log-likelihood:
\begin{align}
    \Ex_{x \sim p(x)}[\log{p(x)}] \ge \Ex_{x \sim p(x)}[\Ex_{z \sim q_\phi(z|x)}[\log{p_\theta(x|z)}] \nonumber\\ - \beta~ \text{D}_{\text{KL}}(q_\phi(z|x)||p(z))],
    \label{obj.betavae}
\end{align}
where setting $\beta=1$ reduces to the original VAE. 
By forcing the variational posterior to be closer to the factorized prior ($\beta > 1$), the model learns a more disentangled representation, but with a sacrifice of generation quality, since it also decreases the mutual information between $z$ and $x$~\cite{dsr/betatcvae,dsr/factorvae}.
To address such trade-off and improve the generation quality, recent approaches propose to gradually anneal the penalty on the KL-divergence~\cite{dsr/annealedvae}, 
or decompose it to isolate the penalty for \textit{total correlation}~\cite{misc/tc} that encourages the statistical independence of latent variables~\cite{dsr/info-dropout,dsr/betatcvae,dsr/factorvae}.

Approaches based on VAE have shown to be effective in learning disentangled representations over a range of tasks from synthetic~\cite{dsr/dsprites} to complex datasets~\cite{dsr/celeba,dsr/chairs}. 
However, their generation performance is generally insufficient to achieve high-fidelity synthesis, even with recent techniques isolating the factorization of the latent variable~\cite{dsr/factorvae, dsr/betatcvae}. 
We argue that this problem is fundamentally attributed to two reasons:
First, most VAE-based approaches assume the fully-independent generative factors~\cite{dsr/betavae, dsr/factorvae, dsr/betatcvae, dsr/google, misc/recentadv, misc/disdis}. 
This strict assumption oversimplifies the latent manifold and may cause the loss of useful information (\eg~correlated factors) for generating realistic data. 
Second, they typically utilize a simple generator, such as the factorized Gaussian decoder, and learn a uni-modal mapping from the latent to input space.
Although this might be useful to learn meaningful representations~\cite{dsr/annealedvae} (\eg~capturing a structure in local modes), such decoder makes it difficult to render complex patterns in outputs (\eg~textures).

\section{High-Fidelity Synthesis via Distillation}
\label{sec:idgan}
\cutsectiondown
Our objective is to build a generative model $G_\omega:\mathcal{Z}\to \mathcal{X}$ that produces high-fidelity output $x\in\mathcal{X}$ with an interpretable latent code $z\in\mathcal{Z}$ (\ie~disentangled representation).
To achieve this goal, we build our framework upon VAE-based models due to their effectiveness in learning disentangled representations. 
However, discussions in the previous section suggest that disentanglement learning in VAE leads to the sacrifice of generation quality due to the strict constraints on fully-factorized latent variables and the utilization of simple decoders.
We aim to improve the VAE-based models by enhancing generation quality while maintaining its disentanglement learning performance.

\begin{figure}[t!]
\centering
\includegraphics[width=0.9\linewidth]{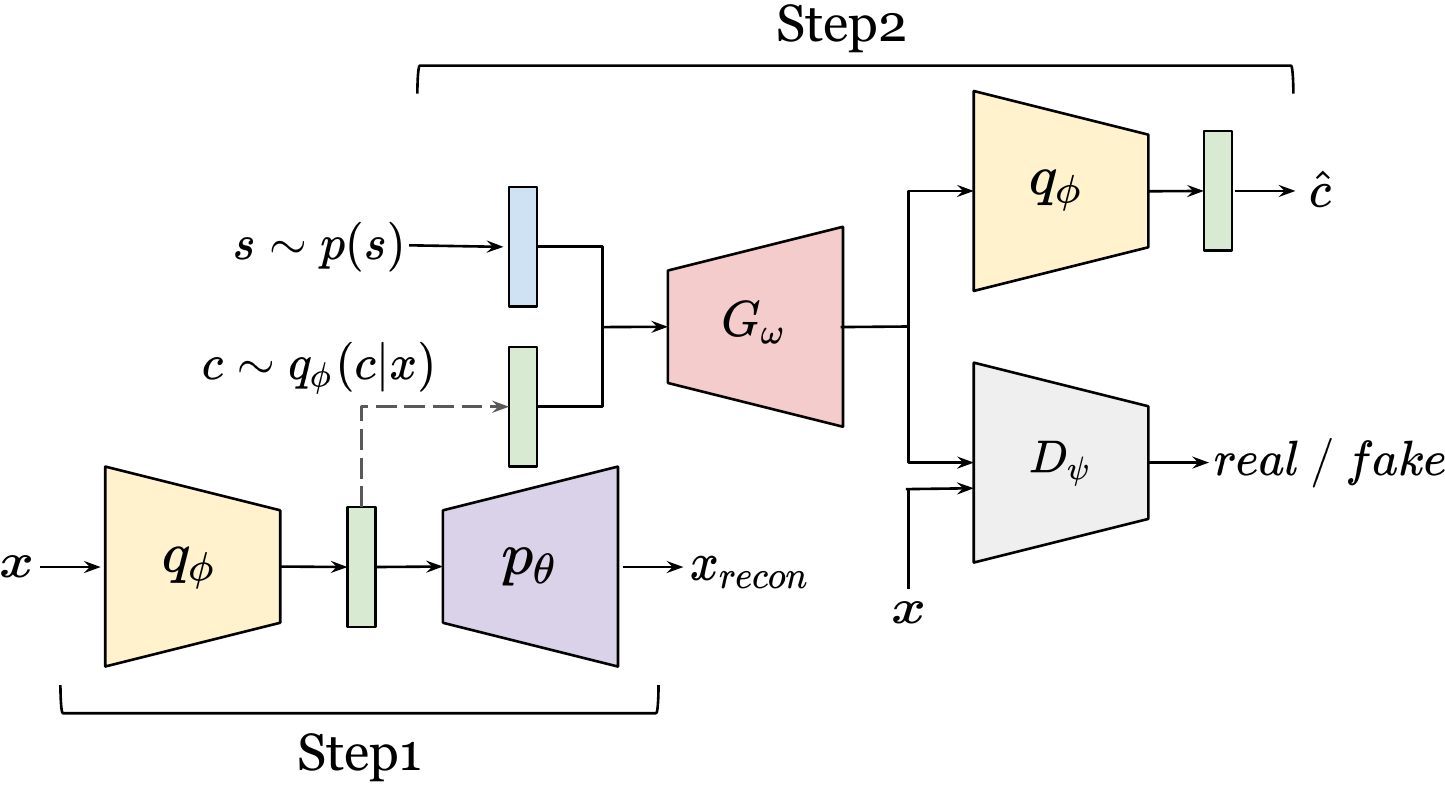}\\
\vspace*{-5pt}
\caption{Overall framework of the proposed method (ID-GAN).}
\label{fig:overview}
\vspace{-0.15in}
\end{figure}

Our main idea is to decompose the objectives of learning disentangled representation and generating realistic outputs into separate but successive learning problems. 
Given a disentangled representation learned by VAEs, we train another network with a much higher modeling capacity (\eg~GAN generator) to decode the learned representation to a realistic sample in the observation space.

Figure~\ref{fig:overview} describes the overall framework of the proposed algorithm.
Formally, let $z=(s,c)$ denote the latent variable composed of the disentangled variable $c$ and the nuisance variable $s$ capturing independent and correlated factors of variation, respectively.
In the proposed framework, we first train VAE (\eg~Eq.~(\ref{obj.betavae})) to learn disentangled latent representations of data, where each observation $x$ can be projected to $c$ by the learned encoder $q_\phi(c|x)$ after the training.
Then in the second stage, we fix the encoder $q_\phi$ and train a generator $G_\omega(z)=G_\omega(s,c)$ for high-fidelity synthesis while \emph{distilling} the learned disentanglement by optimizing the following objective: 
\begin{align}
    &\minl_G \maxl_D~~ \mathcal{L}_{\text{GAN}}(D,G) - \lambda\regours(G),
    \label{eqn:idgan}\\
    &\mathcal{L}_{\text{GAN}}(D,G) = \Ex_{x \sim p(x)}[\log{D(x)}]~+ \label{eqn:GAN_obj}\\
    &~~~~~~~~~~~~~~~~~~~~~~~~~~\Ex_{s \sim p(s), c \sim q_\phi(c)}[\log{(1-D(G(s,c)))}], \nonumber\\
    &\regours(G) = \Ex_{c \sim q_\phi(c), x \sim G(s,c)}[\log{q_\phi(c|x)}]+H_{q_\phi}(c), \label{eqn:infodistill}
\end{align}
where $q_\phi(c)=\frac{1}{N}\sum_{i}q_\phi(c|x_i)$ is the aggregated posterior~\cite{misc/aae, misc/elbo_hoffman, misc/wae} of the encoder network\footnote{In practice, we can easily sample $c$ from $q_\phi(c)$ by $c \sim q_\phi(c|x)p(x)$.}. 
Similar to \cite{dsr/infogan}, Eq.~(\ref{eqn:infodistill}) corresponds to the variational lower-bound of mutual information between the latent code and the generator output $I(c; G(s,c))$, but differs in that (1) $c$ is sampled from the aggregated posterior $q_\phi(c)$ instead of the prior $p(c)$ and (2) it is optimized with respect to the generator only.
Note that we treat $H_{q_\phi}(c)$ as a constant since $q_\phi$ is fixed in Eq.(\ref{eqn:infodistill}).
We refer the proposed model as the Information Distillation Generative Adversarial Network (ID-GAN).

\subsection{Analysis}
\label{sec:anaylsis}
In this section, we provide in-depth analysis of the proposed method and its connections to prior works.

\cutparagraphup
\paragraph{Comparisons to $\beta$-VAEs~\cite{dsr/betavae,dsr/betatcvae,dsr/factorvae}.}%
Despite the simplicity, the proposed ID-GAN effectively addresses the problems in $\beta$-VAEs with generating high-fidelity outputs;
it augments the latent representation by introducing a nuisance variable $s$, which complements the disentangled variable $c$ by modeling richer generative factors.
For instance, the VAE objective tends to favor representational factors that characterize as much data as possible~\cite{dsr/annealedvae} (\eg~azimuth, scale, lighting, \etc.), which are beneficial in representation learning, but incomprehensive to model the complexity of observations. 
Given the disentangled factors discovered by VAEs, the ID-GAN learns to encode the remaining generative factors (such as high-frequency textures, face identity, \etc) into nuisance variable $s$. (Figure~\ref{fig:complex_c_and_z_and_vae}).
This process shares a similar motivation with a progressive augmentation of latent factors~\cite{misc/progressive}, but is used for modeling disentangled and nuisance generative factors.
In addition, ID-GAN employs a much more expressive generator than a simple factorized Gaussian decoder in VAE, which is trained with adversarial loss to render realistic and convincing outputs. 
Combining both, our method allows the generator to synthesize various data in a local neighborhood defined by $c$, where the specific characteristics of each example are fully characterized by the additional nuisance variable $s$.

\cutparagraphup
\paragraph{Comparisons to InfoGAN~\cite{dsr/infogan}.}
The proposed method is closely related to InfoGAN, which optimizes the variational lower-bound of mutual information  $I(c;G(s,c))$ for disentanglement learning.
To clarify the difference between the proposed method and InfoGAN, we rewrite the regularization for both methods using the KL divergence as follows: %
\begin{align}
\reginfo(G,q) &= \Ex_{s \sim p(s)}[D_{\text{KL}}(p(c)||q_\phi(c|G(s,c)))], \label{eqn:infogan_kl}\\
\mathcal{R}_{\text{ours}}(G,q) &= \beta\regvae(q) + \lambda\regours(G),~\text{where} \nonumber\\
\regvae(q) &= \Ex_{x \sim p(x)}[D_{\text{KL}}(q_\phi(c|x)||p(c))], \label{eqn:vae_kl} \\
\regours(G) &= \Ex_{s \sim p(s)}[D_{\text{KL}}(q_\phi(c)||q_\phi(c|G(s,c)))], \label{eqn:idgan_kl} 
\end{align}
where $\mathcal{R}_{\text{ours}}$ summarizes all regularization terms in our method
\footnote{In practice, we learn the encoder $q_\phi$ and generator $G$ independently by Eq.~(\ref{eqn:vae_kl}) and (\ref{eqn:idgan_kl}), respectively, through two-step training.}.
See the Appendix~\ref{supp:kld_derivation} for detailed derivations.

Eq.~(\ref{eqn:infogan_kl}) shows that InfoGAN optimizes the \emph{forward} KL divergence between the prior $p(c)$ and the approximated posterior $q_\phi(c|G(s,c))$.
Due to the zero-avoiding characteristics of forward KL~\cite{misc/forwardkl}, it forces all latent code $c$ with non-zero prior to be covered by the posterior $q_\phi$.
Intuitively, it implies that InfoGAN tries to exploit every dimensions in $c$ to encode each (unique) factor of variations.
It becomes problematic when there is a mismatch between the number of true generative factors and the size of latent variable $c$, which is common in unsupervised disentanglement learning.
On the contrary, VAE optimizes the \emph{reverse} KL divergence (Eq.~(\ref{eqn:vae_kl})), which can effectively avoid the problem by encoding only meaningful factors of variation into certain dimensions in $c$ while collapsing the remainings to the prior.
Since the encoder training in our method is only affected by Eq.~(\ref{eqn:vae_kl}), it allows us to discover the ambient dimension of latent generative factors robust to the choice of latent dimension $|c|$.

In addition, Eq.~(\ref{eqn:infogan_kl}) shows that InfoGAN optimizes the encoder using the generated distributions, which can be problematic when there exists a sufficient discrepancy between the true and generated distributions (\eg~mode-collapse may cause learning partial generative factors.). %
On the other hand, the encoder training in our method is guided by the true data (Eq.~(\ref{eqn:vae_kl})) together with maximum likelihood objective, while the mutual information (Eq.~(\ref{eqn:idgan_kl})) is enforced only to the generator.  This helps our model to discover comprehensive generative factors from data while guiding the generator to align its outputs to the learned representation.

\begin{figure}[t!]
\centering
\includegraphics[width=0.85\linewidth]{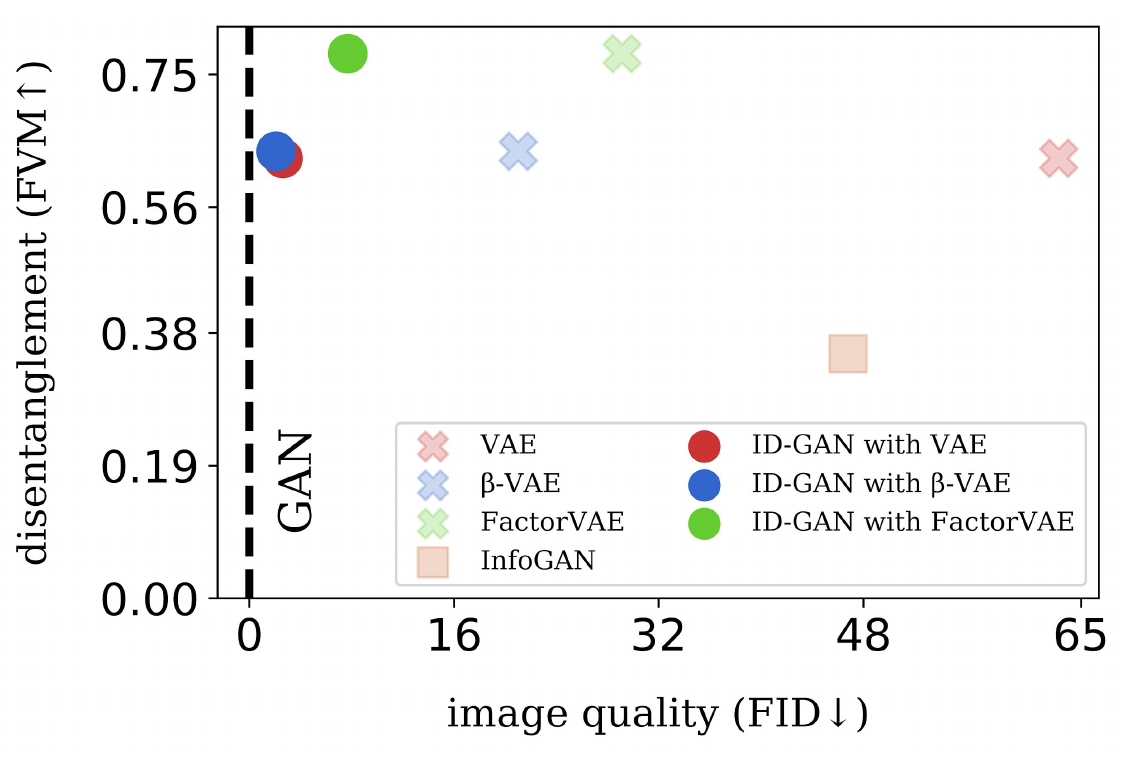}\\
\vspace{-0.05in}
\vspace*{-7pt}
\caption{Comparison of disentanglement vs. generation performance on dSprites dataset.}
\label{fig:dsprites_disent_genqual}
\vspace{-0.15in}
\end{figure}

\cutparagraphup
\paragraph{Practical benefits.}
The objective decomposition in the proposed method also offers a number of practical advantages.
First, it enables plug-and-play-style adoption of the state-of-the-art models for disentangled representation learning and high-quality generation.
As shown in Figure~\ref{fig:dsprites_disent_genqual}, it allows our model to achieve state-of-the-art performance on both tasks.
Second, such decomposition also leads to an efficient model design, where we learn disentanglement from low-resolution images and distill the learned representation to the task of high-resolution synthesis with a much higher-capacity generator.
We argue that it is practically reasonable in many cases since VAEs tend to learn global structures in disentangled representation, which can be captured from low-resolution images.
We demonstrate this in the high-resolution image synthesis task, where we use the disentangled representation learned with $64\times64$ images for the synthesis of $256\times256$ or $1024\times1024$ images.  
\begin{figure*}[!t]
\begin{minipage}{\linewidth}
\centering
\footnotesize
\vspace{-0.1in}
\captionof{table}{
Quantitative comparison results on synthetic datasets. }
\vspace*{-10pt}
\begin{tabular}{l|@{}C{1.2cm}@{}C{1.2cm}@{}C{1.5cm}|@{}C{1.2cm}@{}C{1.2cm}@{}C{1.4cm}|@{}C{1.2cm}@{}C{1.2cm}@{}C{1.4cm}}
\toprule
& \multicolumn{3}{c|}{Color-dSprites} 
& \multicolumn{3}{c|}{Scream-dSprites} 
& \multicolumn{3}{c}{Noisy-dSprites} \\
\hline 
                    & FVM ($\uparrow$)           & MIG ($\uparrow$)           & FID ($\downarrow$)
                    & FVM ($\uparrow$)           & MIG ($\uparrow$)           & FID ($\downarrow$)
                    & FVM ($\uparrow$)           & MIG ($\uparrow$)           & FID ($\downarrow$) \\
\hline
VAE                 & .67$\pm$.12 & .16$\pm$.08 & 21.63$\pm$4.97
                    & .44$\pm$.03 & .08$\pm$.04 & 7.79$\pm$2.51
                    & {\bf .42$\pm$.09} & .05$\pm$.04 & 3.27$\pm$1.94 \\
$\beta$-VAE             & .67$\pm$.07 & .32$\pm$.04 & 15.13$\pm$4.25
                    & {\bf .57$\pm$.01} & {\bf .29$\pm$.00} & 7.33$\pm$2.87
                    & .32$\pm$.05 & .05$\pm$.03	& 3.46$\pm$0.38 \\
FactorVAE           & {\bf .69$\pm$.05} & {\bf .37$\pm$.02} & 10.71$\pm$5.73
                    & {\bf .57$\pm$.01} & .22$\pm$.06 & 6.35$\pm$3.27
                    & .40$\pm$.09 & {\bf .08$\pm$.04} & 2.48$\pm$0.44 \\
\hline
GAN                 & N/A           & N/A           & 0.30$\pm$0.07
                    & N/A           & N/A           & {\bf 0.11$\pm$0.03}
                    & N/A           & N/A           & 9.74$\pm$2.18 \\
InfoGAN             & .34$\pm$ $00$ & .01$\pm$.01  & 30.55$\pm$21.17
                    & .29$\pm$ $00$ & .00$\pm$.00 & 5.77$\pm$3.93
                    & .22$\pm$.02  & .01$\pm$.01 & 5.51$\pm$4.22 \\
\hline
\ours+VAE           & .67$\pm$.12 & .16$\pm$.08 & 0.32$\pm$0.10
                    & .44$\pm$.03 & .08$\pm$.04 & 0.26$\pm$0.03
                    & {\bf .42$\pm$.09} & .05$\pm$.04 & {\bf 1.58$\pm$0.62} \\
\ours+$\beta$-VAE       & .67$\pm$.07 & .32$\pm$.04 & {\bf 0.25$\pm$0.23}
                    & {\bf .57$\pm$.01} & {\bf .29$\pm$.00} & 0.18$\pm$0.02
                    & .32$\pm$.05 & .05$\pm$.03	& 12.42$\pm$1.13 \\
\ours+FactorVAE     & {\bf .69$\pm$.05} & {\bf .37$\pm$.02} & 0.75$\pm$0.54
                    & {\bf .57$\pm$.01} & .22$\pm$.06 & 0.65$\pm$0.33
                    & .40$\pm$.09 & {\bf .08$\pm$.04} & 2.07$\pm$0.87 \\
\bottomrule
\end{tabular}
\label{tab:synthetic_quantitative_eval}
\vspace*{0.15in}
\includegraphics[width=0.75\linewidth]{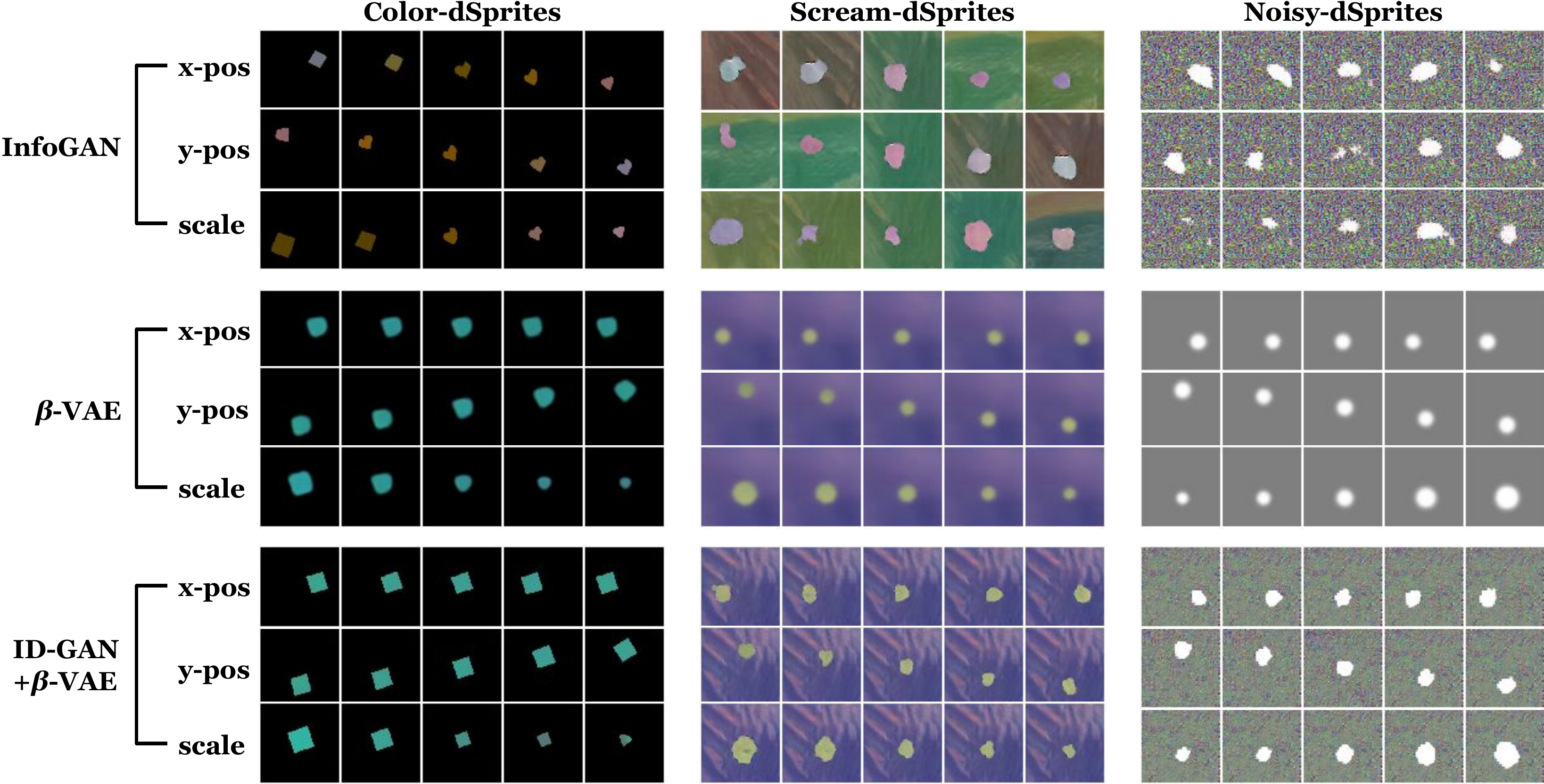}
\captionof{figure}{Qualitative results on synthetic datasets. Both $\beta$-VAE and ID-GAN share the same latent code, but ID-GAN exhibits substantailly higher generation quality.
}
\label{fig:synthetic_qualitative_eval}
\end{minipage}
\vspace{-0.2in}
\vspace*{-9pt}
\end{figure*}
\section{Experiments}
\label{sec:experiments}
\cutsectiondown
In this section, we present various results to show the effectiveness of ID-GAN.
Refer to the Appendix for more comprehensive results and figures.

\subsection{Implementation Details}
\cutsubsectiondown

\paragraph{Compared methods.}
We compare our method with state-of-the-art methods in disentanglement learning and generation.
We choose $\beta$-VAE~\cite{dsr/betavae}, FactorVAE~\cite{dsr/factorvae}, and InfoGAN~\cite{dsr/infogan} as baselines for disentanglement learning.
For fair comparison, we choose the best hyperparameter for each model via extensive hyper-parameter search.
We also report the performance by training each method over five different random seeds and averaging the results.

\cutparagraphup
\paragraph{Network architecture.}
For experiments on synthetic datasets, we adopt the architecture from \cite{dsr/google} for all VAE-based methods (VAE, $\beta$-VAE, and FactorVAE). 
For GAN-based methods (GAN, InfoGAN, and ID-GAN), we employ the same decoder and encoder architectures in VAE as the generator and discriminator, respectively.
We set the size of disentangled latent variable to 10 for all methods, and exclude the nuisance variable in GAN-based methods for a fair comparison with VAE-based methods.
For experiments on complex datasets, we employ the generator and discriminator in the state-of-the-art GAN~\cite{misc/ganbase, misc/ganbasevdb}. 
For VAE architectures, we utilize the same VAE architecture as in the synthetic datasets.
We set the size of disentangled and nuisance variables to 20 and 256, respectively.

\cutparagraphup
\paragraph{Evaluation metrics}
We employ three popular evaluation metrics in the literature: Factor-VAE Metric (FVM)~\cite{dsr/factorvae}, Mutual Information Gap (MIG)~\cite{dsr/betatcvae}, and Fr\'echet Inception Distance (FID)~\cite{misc/fid}.
\emph{FVM} and \emph{MIG} evaluate the disentanglement performance by measuring the degree of axis-alignment between each dimension of learned representations and ground-truth factors.
\emph{FID} evaluates the generation quality by measuring the distance between the true and the generated distributions.

\subsection{Results on Synthetic Dataset.}
\label{sec:synthetic_experiments}
\cutsubsectiondown

For quantitative evaluation of disentanglement, we employ the dSprites dataset~\cite{dsr/dsprites}, which contains synthetic images generated by randomly sampling known generative factors, such as shape, orientation, size, and x-y position.
Since the complexity of dSprites is limited to analyze the disentanglement and generation performance, we adopt three variants of dSprites, which are generated by adding color~\cite{dsr/factorvae} (Color-dSprites) or background noise~\cite{dsr/google} (Noisy- and Scream-dSprites).

Table~\ref{tab:synthetic_quantitative_eval} and Figure~\ref{fig:synthetic_qualitative_eval} summarize the quantitative and qualitative comparison results with existing disentanglement learning approaches, respectively.
First, we observe that VAE-based approaches (\ie~$\beta$-VAE and FactorVAE) achieve the state-of-the-art disentanglement performance across all datasets, outperforming the VAE baseline and InfoGAN with a non-trivial margin.
The qualitative results in Figure~\ref{fig:synthetic_qualitative_eval} show that the learned generative factors are well-correlated with meaningful disentanglement in the observation space.
On the other hand, InfoGAN fails to discover meaningful disentanglement in most datasets.
We observe that information maximization in InfoGAN often leads to undesirable factorization of generative factors,
such as encoding both shape and position into one latent code, but factorizing latent dimensions by different combinations of them (\eg~Color-dSprites in Figure~\ref{fig:synthetic_qualitative_eval}).
ID-GAN achieves state-of-the-art disentanglement through the distillation of the learned latent code from the VAE-based models.
Appendix~\ref{supp:sensitivity} also shows that ID-GAN is much more stable to train and insensitive to hyper-parameters than InfoGAN.

In terms of generation quality, VAE-based approaches generally perform much worse than GAN baseline.
This performance gap is attributed to the strong constraints on the factorized latent variable and weak decoder in VAE, which limits the generation capacity.
This is clearly observed in the results on the Noisy-dSprites dataset (Figure~\ref{fig:synthetic_qualitative_eval}), where the outputs from $\beta$-VAE fail to render the high-dimensional patterns in the data (\ie~uniform noise).
On the other hand, our method achieves competitive generation performance to the state-of-the-art GAN using a much more flexible generator for synthesis, which enables the modeling of complex patterns in data.
As observed in Figure~\ref{fig:synthetic_qualitative_eval}, ID-GAN performs generation using the \emph{same} latent code with $\beta$-VAE, but produces much more realistic outputs by capturing accurate object shapes (in Color-dSprites) and background patterns (in Scream-dSprites and Noisy-dSprites) missed by the VAE decoder.
These results suggest that our method can achieve the best trade-off between disentanglement learning and high-fidelity synthesis.

\cutsubsectionup
\subsection{Ablation Study}
\label{sec:ablation_study}
\cutsubsectiondown
This section provides an in-depth analysis of our method.  

\begin{table}
\footnotesize
\caption{
Comparison of approaches using a joint and decomposed objective for disentanglement learning and synthesis. %
}
\vspace*{-10pt}
\begin{tabular}{l|@{}C{1.7cm}@{}C{1.7cm}@{}C{1.7cm}}
\toprule
& \multicolumn{3}{c}{dSprites} \\
\hline 
                            & FVM ($\uparrow$)      & MIG ($\uparrow$)      & FID ($\downarrow$) \\
\hline
$\beta$-VAE (reference)                 & \bf{0.65$\pm$0.08}           & \bf{0.28$\pm$0.09}           & 37.75$\pm$24.58 \\
\hline
VAE-GAN                     & 0.46$\pm$0.18           & 0.13$\pm$0.11           & 33.54$\pm$24.93 \\
ID-GAN (end-to-end)         & 0.50$\pm$0.14                 & 0.13$\pm$0.09                 & 3.18$\pm$2.38 \\
\hline
ID-GAN (two-step)           & \bf{0.65$\pm$0.08}                 & \bf{0.28$\pm$0.09}                 & \bf{2.00$\pm$1.74} \\
\bottomrule
\end{tabular}
\label{tab:synthetic_ablative_eval}
\vspace{-0.1in}
\end{table}
\cutparagraphup
\paragraph{Is two-step training necessary?}
\label{sec:ablation_twostep}
First, we study the impact of two-stage training for representation learning and synthesis. 
We consider two baselines: (1) VAE-GAN~\cite{misc/vaegan} as an extension of $\beta$-VAE with adversarial loss, and (2) end-to-end training of ID-GAN.
Contrary to ID-GAN that learns to represent ($q_\phi$) and synthesize ($G$) data via separate objectives, these baselines learn a single, entangled objective for both tasks.
Table \ref{tab:synthetic_ablative_eval} summarizes the results in the dSprites dataset.

The results show that VAE-GAN improves the generation quality of $\beta$-VAE with adversarial learning.
The generation quality is further improved in the end-to-end version of ID-GAN by employing a separate generator for synthesis.  
However, the improved generation quality in both baselines comes with the cost of degraded disentanglement performance.
We observe that updating the encoder using adversarial loss hinders the discovery of disentangled factors, as the discriminator tends to exploit high-frequency details to distinguish the real images from the fake images, which motivates the encoder to learn nuisance factors.
This suggests that decomposing the representation learning and generation objective is important in the proposed framework (ID-GAN two-step), which achieves the best performance in both tasks.
\begin{table}
\begin{minipage}[b]{\linewidth}
\caption{
Comparison of two-step approaches for generation (FID) and alignment ($\regours$ and GILBO)\protect\footnotemark performance.}
\vspace*{-10pt}
\footnotesize
\label{tab:complex_ablative_eval}
\centering
\begin{tabular}{l|@{}C{1.7cm}@{}C{1.7cm}@{}C{1.7cm}}
\toprule
& \multicolumn{3}{c}{CelebA 128x128} \\
\hline 
                            & FID ($\downarrow$)    & {$\regours$} ($\uparrow$) & GILBO~\cite{gilbo} ($\uparrow$) \\
\hline
ID-GAN w/o distill          &   \textbf{5.75}       &    -65.84            & -20.40  \\
cGAN                        &   7.07       &    -17.39            & -7.57  \\
\hline
ID-GAN                      &   6.61       &    \textbf{-10.25}            & \textbf{-0.19}   \\
\bottomrule
\end{tabular}
\centering
\includegraphics[width=0.9\linewidth]{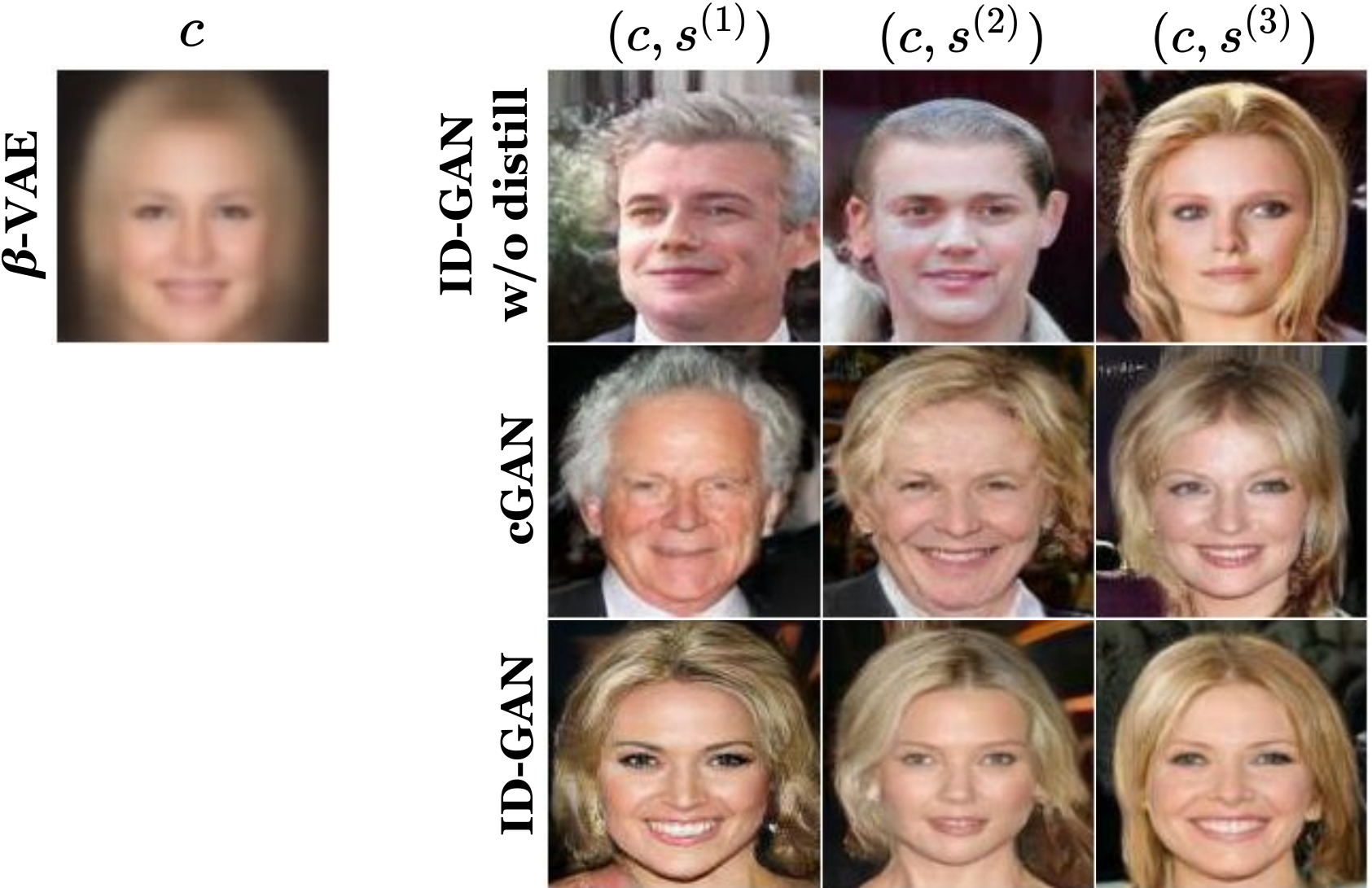}
\captionof{figure}{
Qualitative comparisons of various two-step approaches.
All samples share the same disentangled code $c$, but different nuisance variable $s$.
(1) First column: output of $\beta$-VAE decoder.
(2) Second to fourth columns: images generated by different nuisance variables $s$ using various methods (rows). 
}
\label{fig:idloss_effect}
\vspace{-0.1in}
\vspace*{-10pt}
\end{minipage}
\end{table}\footnotetext{We report both $\regours$ and GILBO without $H_{q_\phi}(c)$ to avoid potential error in measuring $q_\phi(c)$ (\eg~fitting a Gaussian~\cite{gilbo}). Note that it does not affect the relative comparison since all models share the same $q_\phi$.}

\cutparagraphup
\paragraph{Is distillation necessary?}
\label{sec:ablation_distillation}
The above ablation study justifies the importance of two-step training. 
Next, we compare different approaches for two-step training that perform conditional generation using the representation learned by $\beta$-VAE.
Specifically, we consider two baselines:
(1) cGAN and (2) ID-GAN trained without distillation (ID-GAN w/o distill).
We opt to consider cGAN as the baseline since we find that it implicitly optimizes $\regours$ (see Appendix~\ref{supp:sec:cgan_proof} for the proof).
In the experiments, we train all models in the CelebA 128x128 dataset using the same $\beta$-VAE trained on the $64\times64$ resolution, and compare the generation quality (FID) and a degree of alignment between the disentangled code $c$ and generator output $G(s,c)$.
For comparison of the alignment, we measure $\regours$ (Eq.~(\ref{eqn:idgan_kl})) and GILBO
\footnote{GILBO is formulated similarly as $\regours$ (Eq.~(\ref{eqn:infodistill})), but optimized over another auxiliary encoder network different from the one used in $\regours$.} 
\cite{gilbo}, both of which are valid lower-bounds of mutual information $I(c; G(s,c))$.
Note that the comparison based on the lower-bound is still valid as its relative order has shown to be insensitive to the tightness of the bound~\cite{gilbo}.   
Table~\ref{tab:complex_ablative_eval} and Figure~\ref{fig:idloss_effect} summarize the quantitative and qualitative results, respectively.

As shown in the table, all three models achieve comparable generation performances in terms of FID.
However, we observe that their alignments to the input latent code vary across the methods.
For instance, ID-GAN (w/o distill) achieves very low $\regours$ and GILBO,
indicating that the generator output is not accurately reflecting the generative factors in $c$. 
The qualitative results (Figure~\ref{fig:idloss_effect}) also show considerable mismatch between the $c$ and the generated images.
Compared to this, cGAN achieves much higher degree of alignment due to the implicit optimization of $\regours$, but its association is much loose than our method (\eg~changes in gender and hairstyle).
By explicitly constraining the generator to optimize $\regours$, ID-GAN achieves the best alignment. 

\cutsubsectionup
\subsection{Results on Complex Dataset}
\cutsubsectiondown

To evaluate our method with more diverse and complex factors of variation, we conduct experiments on natural image datasets, such as CelebA~\cite{dsr/celeba}, 3D Chairs~\cite{dsr/chairs}, and Cars~\cite{dsr/cars}.
We first evaluate our method on $64\times64$ images, and extend it to higher resolution images using the CelebA ($256\times256$) and CelebA-HQ~\cite{dsr/celeba-hq} ($1024\times1024$) datasets.

\cutparagraphup
\paragraph{Comparisons to other methods.}
Table~\ref{tab:complex_quantitative_eval} summarizes quantitative comparison results (see Appendix~\ref{supp:fig:idgan_complexc_samples} for qualitative comparisons). 
Since there are no ground-truth factors available in these datasets, we report the performance based on generation quality (FID).
As expected, the generation quality of VAE-based methods is much worse in natural images. 
GAN-based methods, on the contrary, can generate more convincing samples exploiting the expressive generator.
However, we observe that the baseline GAN taking only nuisance variables ends up learning highly-entangled generative factors.
ID-GAN achieves disentanglement via disentangled factors learned by VAE, and generation performance on par with the GAN baseline.

\begin{figure}[!t]
     \centering
     \includegraphics[width=1\linewidth]{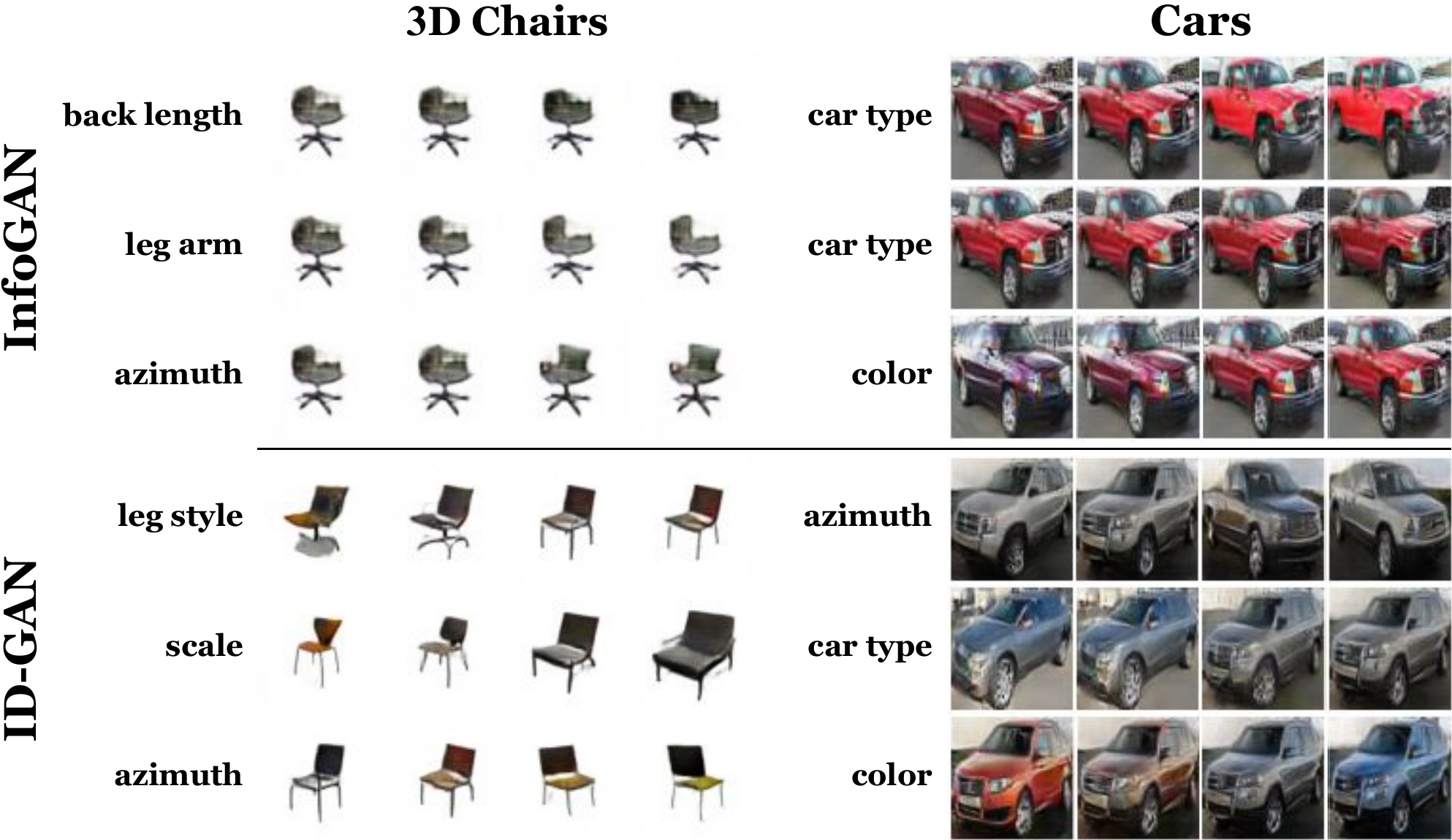}
     \vspace*{-20pt}
     \caption{
     Comparisons of latent traversal between GAN-based approaches. Although both methods achieve comparable generation quality, ID-GAN learns much more meaningful disentanglement. 
     }
     \vspace*{-5pt}
     \label{fig:complex_gan_traversal}
     \vspace{-0.1in}
\end{figure}

To better understand the disentanglement learned by GAN-based methods, we present latent traversal results in Figure~\ref{fig:complex_gan_traversal}.
We generate samples by modifying values of each dimension in the disentangled latent code $c$ while fixing the rest.
We observe that the InfoGAN fails to encode meaningful factors into $c$, and nuisance variable $z$ dominates the generation process, making all generated images almost identical.
On the other hand, ID-GAN learns meaningful disentanglement with $c$ and generates reasonable variations.

\begin{table}[!t]
\centering
\footnotesize
\caption{Quantitative results based on FID ($\downarrow$).}
\vspace*{-10pt}
\begin{tabular}{l|@{}C{1.2cm}@{}C{1.2cm}@{}C{1.2cm}}
\toprule
& 3D chair & Cars & CelebA\\
\hline
VAE                 & 116.46 &	201.29 & 160.06 \\
                    
$\beta$VAE             & 107.97 &	235.32 & 166.01 \\
                    
FactorVAE           & 123.64 &  208.60 & 154.48 \\

\hline
GAN                 & {\bf 24.17}	& 14.62	& {\bf 3.34} \\
                    
InfoGAN             & 60.45	& {\bf 13.67} & 4.93 \\
\hline
\ours+$\beta$VAE       & 25.44 & 14.96	& 4.08 \\
\bottomrule
\end{tabular}
\label{tab:complex_quantitative_eval}
\vspace{-0.1in}
\end{table}

 \begin{figure}[!t]
    \centering
        \includegraphics[width=1\linewidth]{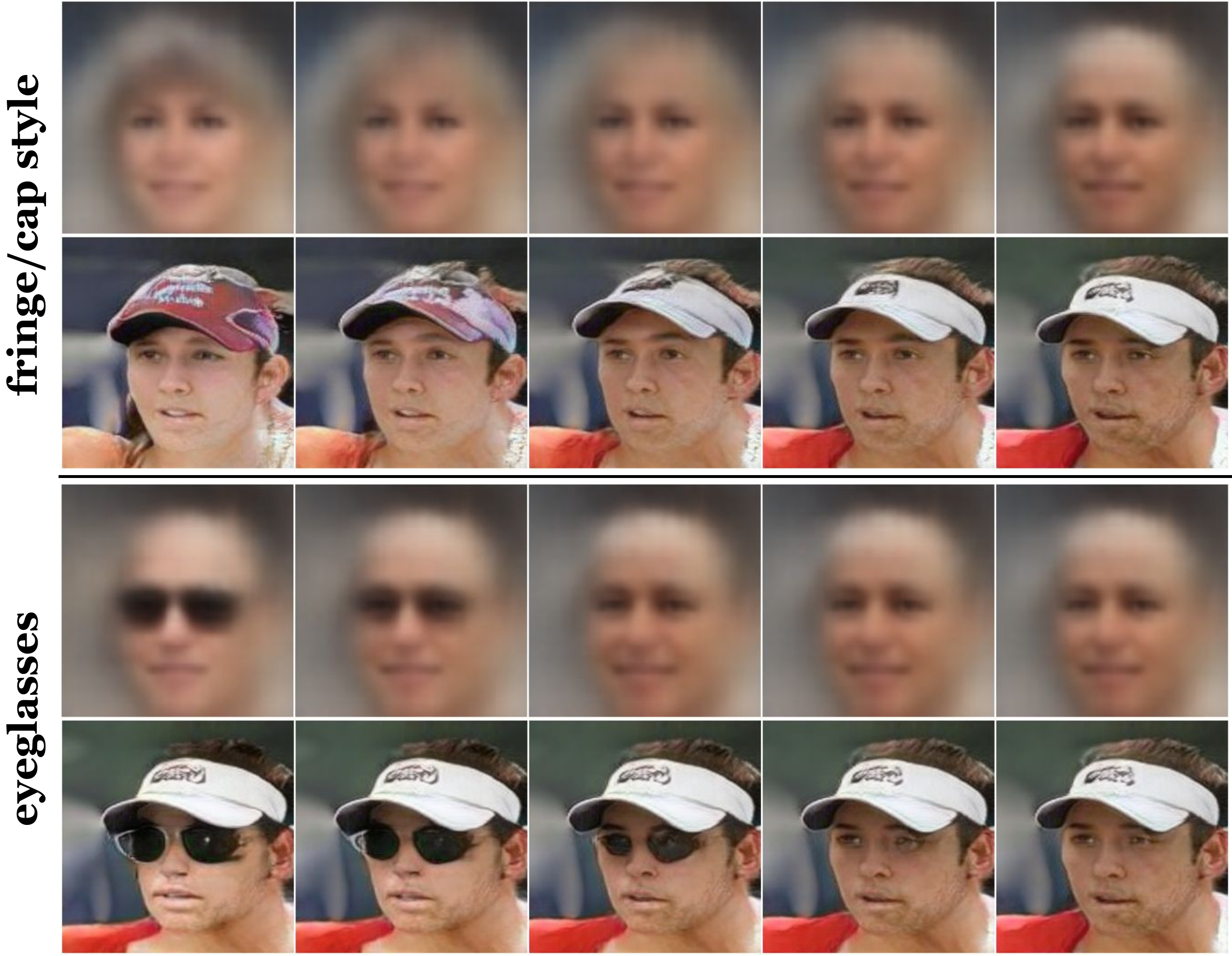}
    \vspace*{-20pt}
    \caption{Comparisons of VAE and ID-GAN outputs (top-rows: VAE, bottom-rows: ID-GAN). 
    Note that both outputs are generated from the same latent code, but using different decoders. 
    Both decoders are aligned well to render the same generative factors, but ID-GAN produces much more realistic outputs.
    }
    \label{fig:complex_latent_traversal}
    \vspace{-0.1in}
\end{figure}

\cutparagraphup
\vspace{-0.03in}
\paragraph{Extension to high-resolution synthesis.}
One practical benefit of the proposed two-step approach is that we can incorporate any VAE and GAN into our framework. 
To demonstrate this, we train ID-GAN for high-resolution images (\eg~$256\times256$ and $1024\times1024$) while distilling the $\beta$-VAE encoder learned with \emph{much smaller} $64\times 64$ images\footnote{We simply downsample the generator output by bilinear sampling to match the dimension between the generator and encoder.}.
This allows us to easily scale up the resolution of synthesis and helps us to better assess the disentangled factors.

\begin{figure}[!t]
    \centering
    \includegraphics[width=1\linewidth]{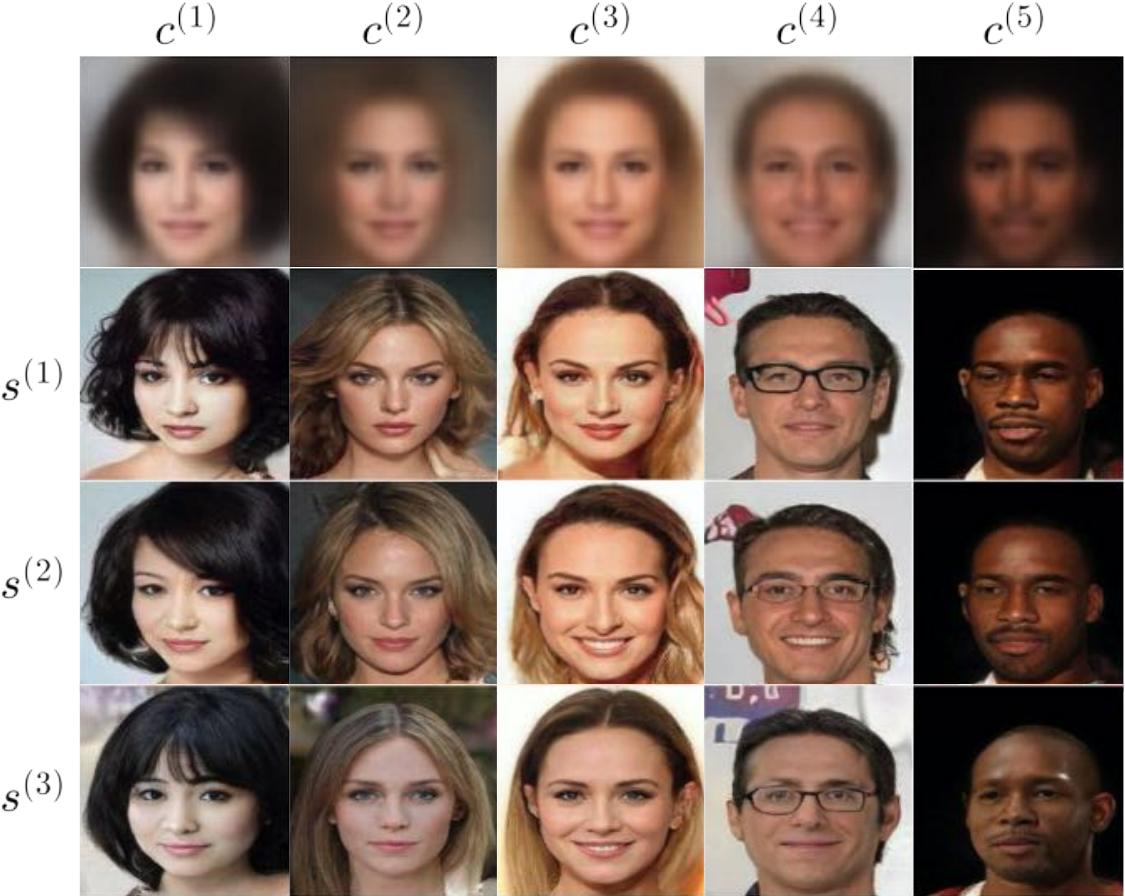}
    \vspace*{-20pt}
    \caption{Analysis on the learned disentangled variables $c^{(m)} \in \mathbb{R}^{20}$ and nuisance variables $s^{(n)}\in \mathbb{R}^{256}$ of ID-GAN on CelebA (256$\times$256). The samples in the first row are generated by the $\beta$-VAE decoder and the rest are generated by ID-GAN. Each $c^{(m)}$ captures the most salient factors of variation (\eg., azimuth, hair-structure, \etc.) while $s^{(n)}$ contributes to the local details (\eg., $s^{(2)}$ and $s^{(3)}$ for curvy and straight hair, respectively). %
    }
    \label{fig:complex_c_and_z_and_vae}
    \vspace{-0.15in}
\end{figure}

We first adapt ID-GAN to the $256\times256$ image synthesis task.
To understand the impact of distillation, we visualize the outputs from the VAE decoder and the GAN generator using the same latent code as inputs.  
Figure~\ref{fig:complex_latent_traversal} summarizes the results.
We observe that the outputs from both networks are aligned well to render the same generative factors to similar outputs.
Contrary to blurry and low-resolution ($64\times64$) VAE outputs, however, ID-GAN produces much more realistic and convincing outputs by introducing a nuisance variable and employing more expressive decoder trained on higher-resolution ($256\times256$).
Interestingly, synthesized images by ID-GAN further clarify the disentangled factors learned by the VAE encoder.
For instance, the first row in Figure~\ref{fig:complex_latent_traversal} shows that the ambiguous disentangled factors from the VAE decoder output is clarified by ID-GAN, which is turned out to capture the style of a cap.  
This suggests that ID-GAN can be useful in assessing the quality of the learned representation, which will broadly benefit future studies. 

To gain further insights on the learned generative factors by our method, we conduct qualitative analysis on the latent variables ($c$ and $s$)
by generating samples by fixing one variable while varying another (Figure~\ref{fig:complex_c_and_z_and_vae}).
We observe that varying the disentangled variable $c$ leads to variations in the holistic structures in the outputs, such as azimuth, skin color, hair style, etc, while varying the nuisance variable $s$ leads to changes in more fine-grained facial attributes, such as expression, skin texture, identity, \etc.
It shows that ID-GAN successfully distills meaningful and representative disentangled generative factors learned by the inference network in VAE, while producing diverse and high-fidelity outputs using generative factors encoded in the nuisance variable.

\begin{figure}[t]
\centering
\includegraphics[width=\linewidth]{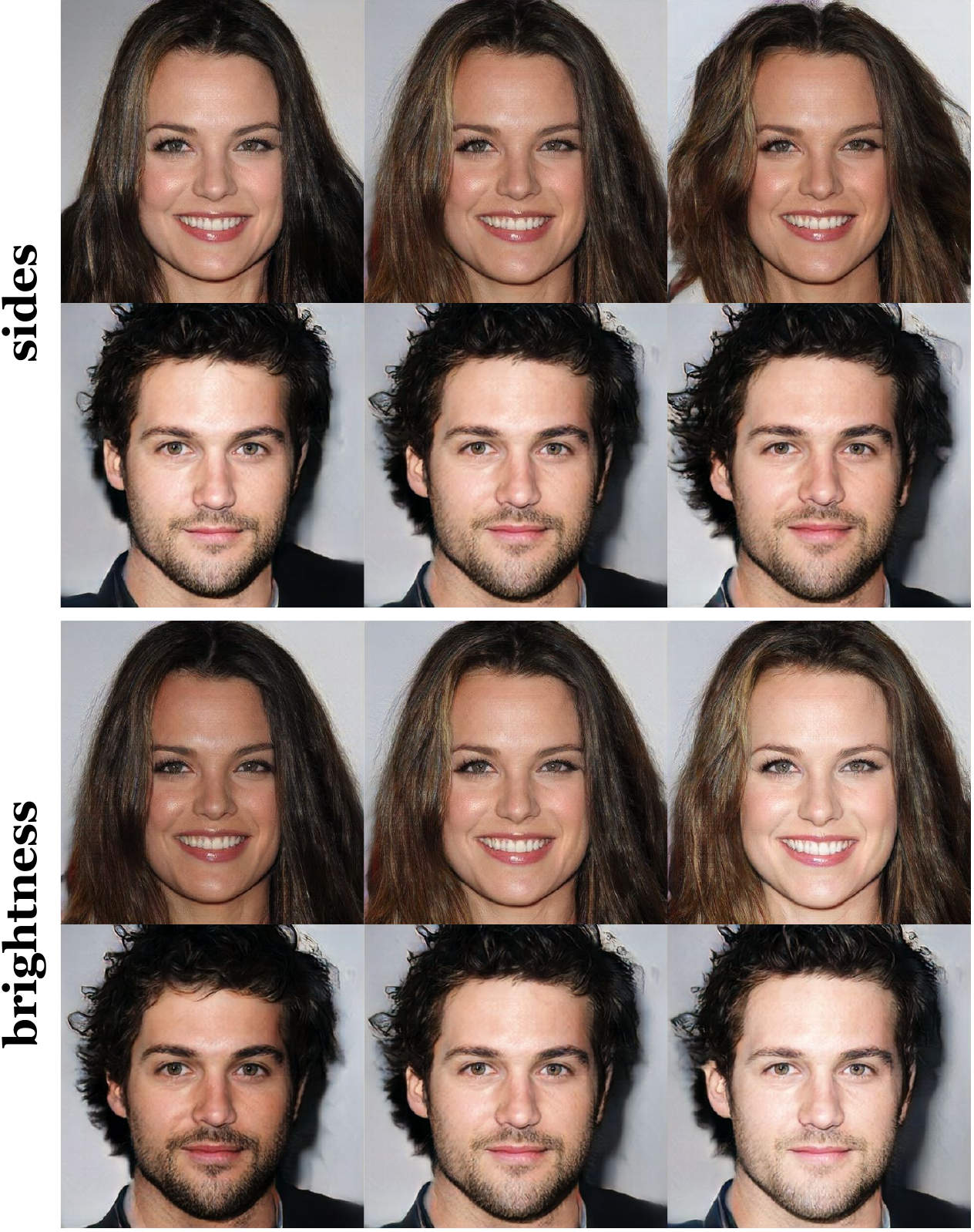}
\vspace*{-20pt}
\caption{Results on the CelebA-HQ dataset ($1024\times1024$ images).}
\label{fig:complex_celebAHQ}
\vspace{-0.15in}
\end{figure}

Finally, we further conduct experiments on the more challenging task of mega-pixel image synthesis.
In the experiments, we base our ID-GAN on the VGAN architecture~\cite{misc/ganbasevdb} and adapt it to synthesize CelebA-HQ $1024\times1024$ images  given factors learned by $\beta$-VAE.
Figure~\ref{fig:complex_celebAHQ} presents the results, where we generate images by changing one values in one latent dimension in $c$.
We observe that ID-GAN produces high-quality images with nice disentanglement property, where it changes one factor of variation in the data (\eg~azimuth and hair-style) while preserving the others (\eg~identity).

\section{Conclusion}
\label{sec:conclusion}
We propose Information Distillation Generative Adversarial Network (ID-GAN), a simple framework that combines the benefits of the disentanglement representation learning and high-fidelity synthesis.
We show that we can incorporate the state-of-the-art for both tasks by decomposing their objectives while constraining the generator by distilling the encoder.  
Extensive experiments on synthetic and complex datasets validate that the proposed method can achieve the best trade-off between realism and disentanglement, outperforming the existing approaches with substantial margin.
We also show that such decomposition leads to efficient and effective model design, allowing high-fidelity synthesis with disentanglement on high-resolution images.

{\small
\bibliographystyle{ieee_fullname}
\bibliography{dsr}
}
\clearpage
\captionsetup[subfigure]{labelformat=parens,labelsep=space,font=small}
\captionsetup[figure]{labelformat=simple,labelsep=period,font=normalsize}
\captionsetup[table]{labelformat=simple,labelsep=period,font=normalsize}

\renewcommand{\thefigure}{A.\arabic{figure}}
\renewcommand{\theequation}{A.\arabic{equation}}
\renewcommand{\thetable}{A.\arabic{table}}
\setcounter{figure}{0}
\setcounter{equation}{0}
\setcounter{table}{0}

\onecolumn
\appendix
\section*{Appendix}
\addcontentsline{toc}{section}{Appendix}
\renewcommand{\thesection}{\Alph{section}}

\section{Derivations}
\subsection{InfoGAN optimizes Forward-KL Divergence}
\label{supp:kld_derivation}
This section provides the derivation of Eq.~(\ref{eqn:infogan_kl}) in the main paper.
Consider $G(s,c)$ as a mapping function $G: \mathcal{S \times C} \rightarrow \mathcal{X}$, where $s$ and $c$ denote nuisance and disentangled variables, respectively.
Also, assuming $G$ is a deterministic function of $(s, c)$, the conditional distribution $p_G(x|s,c)$ can be approximated to a dirac distribution $\delta(x-G(s,c))=\mathbbm{1}\{x=G(s,c)\}$.
Then, the marginal distribution $p_G(x)$ can be described as below:
\begin{align}
    p_G(x) =& ~\int_{s}\int_{c}{p(s)p(c)p_G(x|s,c)dcds} = E_{s \sim p(s), c \sim p(c)}[p_G(x|s,c)] = E_{s \sim p(s), c \sim p(c)}[\mathbbm{1}\{x=G(s,c)\}].
\end{align}
Then, the variational lower-bound of mutual information optimized in InfoGAN~\cite{dsr/infogan} can be rewritten as follows:
\begin{align}
    \reginfo(G,q) =& ~E_{c \sim p(c), x \sim G(s,c)}[\log{q_\phi(c|x)}] + H(c), \nonumber \\
                =& ~\int_{s}p(s)\int_{c}p(c)\int_{x}p_G(x|s,c)\log{q_\phi(c|x)}dxdcds + H(c), \\
                =& ~\int_{s}p(s)\int_{c}p(c)\int_{x}\mathbbm{1}\{x=G(s,c)\}\log{q_\phi(c|x)}dxdcds + H(c), \\
                =& ~\int_{s}p(s)\int_{c}p(c)\log{q_\phi(c|G(s,c))}dcds + H(c), \\
                =& ~\int_{s}p(s)\int_{c}p(c)\log{q_\phi(c|G(s,c))}dcds - \int_{c}p(c)\log{p(c)}dc, \\
                =& ~\int_{s}p(s)\int_{c}p(c)\log{q_\phi(c|G(s,c))}dcds - \int_{c}p(c)\log{p(c)}dc\int_{s}p(s)ds, \\
                =& ~\int_{s}p(s)\int_{c}p(c)\log{\frac{q_\phi(c|G(s,c))}{p(c)}}dcds, \\
                =& ~-E_{s \sim p(s)}[D_{KL}(p(c)||q_\phi(c|G(s,c))],\label{eqn:supp_infokl}
\end{align}
where the Eq.~(\ref{eqn:supp_infokl}) corresponds to Eq.~(\ref{eqn:infogan_kl}) of the main paper. Similiarly, we can rewrite the distillation regularization $\regours(G)$ (Eq.~(\ref{eqn:infodistill}) in the main paper) as follows:
\begin{align}
    \regours(G) =& ~E_{c \sim q_\phi(c), x \sim G(s,c)}[\log{q_\phi(c|x)}] + H_{q_\phi}(c), \nonumber \\
                =& ~\int_{s}p(s)\int_{c}q_\phi(c)\int_{x}\mathbbm{1}\{x=G(s,c)\}\log{q_\phi(c|x)}dxdcds + H_{q_\phi}(c), \\
                =& ~\int_{s}p(s)\int_{c}q_\phi(c)\log{q_\phi(c|G(s,c))}dcds - \int_{s}p(s)\int_{c}q_\phi(c)\log{q_\phi(c)}dcds, \\
                =& ~-E_{s \sim p(s)}[D_{KL}(q_\phi(c)||q_\phi(c|G(s,c))],\label{eqn:supp_ourskl}
\end{align}
where Eq. (\ref{eqn:supp_ourskl}) corresponds to Eq. (\ref{eqn:idgan_kl}) in the main paper.
As discussed in the paper, both Eq. (\ref{eqn:supp_infokl}) and (\ref{eqn:supp_ourskl}) correspond to the \emph{forward} KLD;
regularization in InfoGAN $\reginfo(G,q)$ (Eq. (\ref{eqn:supp_infokl})) is optimized with respect to both the encoder $q$ and the generator $G$, which is problematic due to the zero-avoiding characteristics of forward KLD and the potential mismatch between the true and generated data distributions.
On the other hand, our method can effectively avoid this problem by optimizing Eq. (\ref{eqn:supp_ourskl}) with only respect to the generator while encoder training is guided by \emph{reverse} KLD using the true data distribution (Eq. (\ref{eqn:vae_kl}) in the main paper).

\subsection{cGAN implicitly maximizes $\regours(G)$}
\label{supp:sec:cgan_proof}
In Section \ref{sec:ablation_study} of the main paper, we define cGAN as the baseline that also optimizes $\regours(G)$ implicitly in its objective function.
This section provides its detailed derivation. 
Formally, we consider cGAN that minimizes a Jensen-Shannon divergence (JSD) between two joint distributions $\JSD(p_d(x,c)||p_G(x,c))$, where $p_d(x,c)=p(x)q_\phi(c|x)$ and $p_G(x,c)=q_\phi(c)p_G(x|c)=q_\phi(c)\int_s{p(s)p_G(x|s,c)ds}$ denote real and fake joint distributions, respectively.
Then, $\regours(G)$ from $\JSD(p_d(x,c)||p_g(x,c))$ is derived as follows:
\begin{align}
    2\JSD(p_d(x,c)||p_g(x,c)) 
       =& ~\KLD(p_d(x,c)||p_G(x,c)) + \KLD(p_G(x,c)||p_d(x,c)), \\
       =& ~\KLD(p_d(x,c)||p_G(x,c)) + \int_{c,x}p_G(x,c)\log{\frac{p_G(x,c)}{p_d(x,c)}}dxdc, \\
       =& ~\KLD(p_d(x,c)||p_G(x,c)) + \int_{c,x}q_\phi(c)p_G(x|c)\log{\frac{q_\phi(c)p_G(x|c)}{p(x)q_\phi(c|x)}}dxdc, \\
       =& ~\KLD(p_d(x,c)||p_G(x,c)), \nonumber \\
       &+ \int_{c,x}q_\phi(c)p_G(x|c)\log{\frac{p_G(x|c)}{p(x)}}dxdc + \int_{c,x}q_\phi(c)p_G(x|c)\log{\frac{q_\phi(c)}{q_\phi(c|x)}}dxdc,\\
       =& ~\KLD(p_d(x,c)||p_G(x,c)), \nonumber \\
       &+ \int_{c}q_\phi(c)\int_{x}p_G(x|c)\log{\frac{p_G(x|c)}{p(x)}}dxdc + \int_{s}p(s)\int_{c}q_\phi(c)\int_{x}p_G(x|s,c)\log{\frac{q_\phi(c)}{q_\phi(c|x)}}dxdcds, \\
       =& ~\KLD(p_d(x,c)||p_G(x,c)) + \mathbb{E}_{c \sim q_\phi(c)}[\KLD(p_G(x|c)||p(x)] - \regours(G). \label{eqn:cgan_idloss}
\end{align}
Eq.~(\ref{eqn:cgan_idloss}) implies that the cGAN objective also implicitly maximizes $\regours(G)$.
However, Eq.~(\ref{eqn:cgan_idloss}) is guaranteed only when the discriminator converges to (near-)optimal with respect to real and fake joint distributions, which makes the optimization of $\regours(G)$ highly dependent on the quality of the discriminator.
On the other hand, ID-GAN maximizes $\regours(G)$ explicitly by directly computing $\regours(G)$ from the learned encoder $q_\phi$, which leads to a higher degree of alignment between the input latent code and the generated output (Table \ref{tab:complex_ablative_eval} and Figure \ref{fig:idloss_effect} in the main paper).

\clearpage

\section{Additional Experiment Results}
\subsection{Additional Results on Synthetic Dataset}
\begin{figure*}[htbp!]
    \centering
    \begin{subfigure}[b]{0.38\textwidth}
        \centering
        \includegraphics[width=\textwidth]{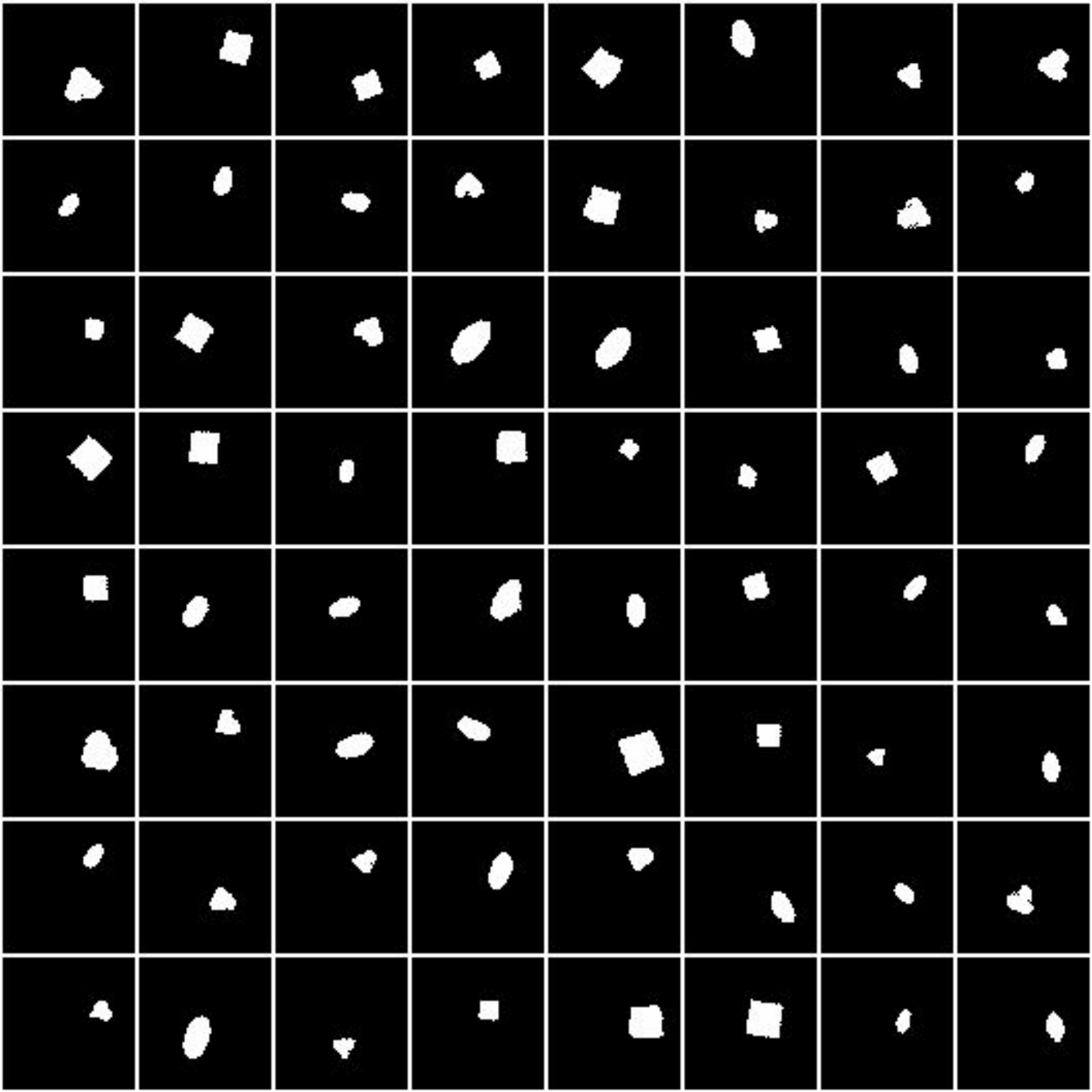}
        \caption[]
        {{\small dSprites}}
    \end{subfigure}
    \hspace{0.3in}
    \begin{subfigure}[b]{0.38\textwidth}  
        \centering 
        \includegraphics[width=\textwidth]{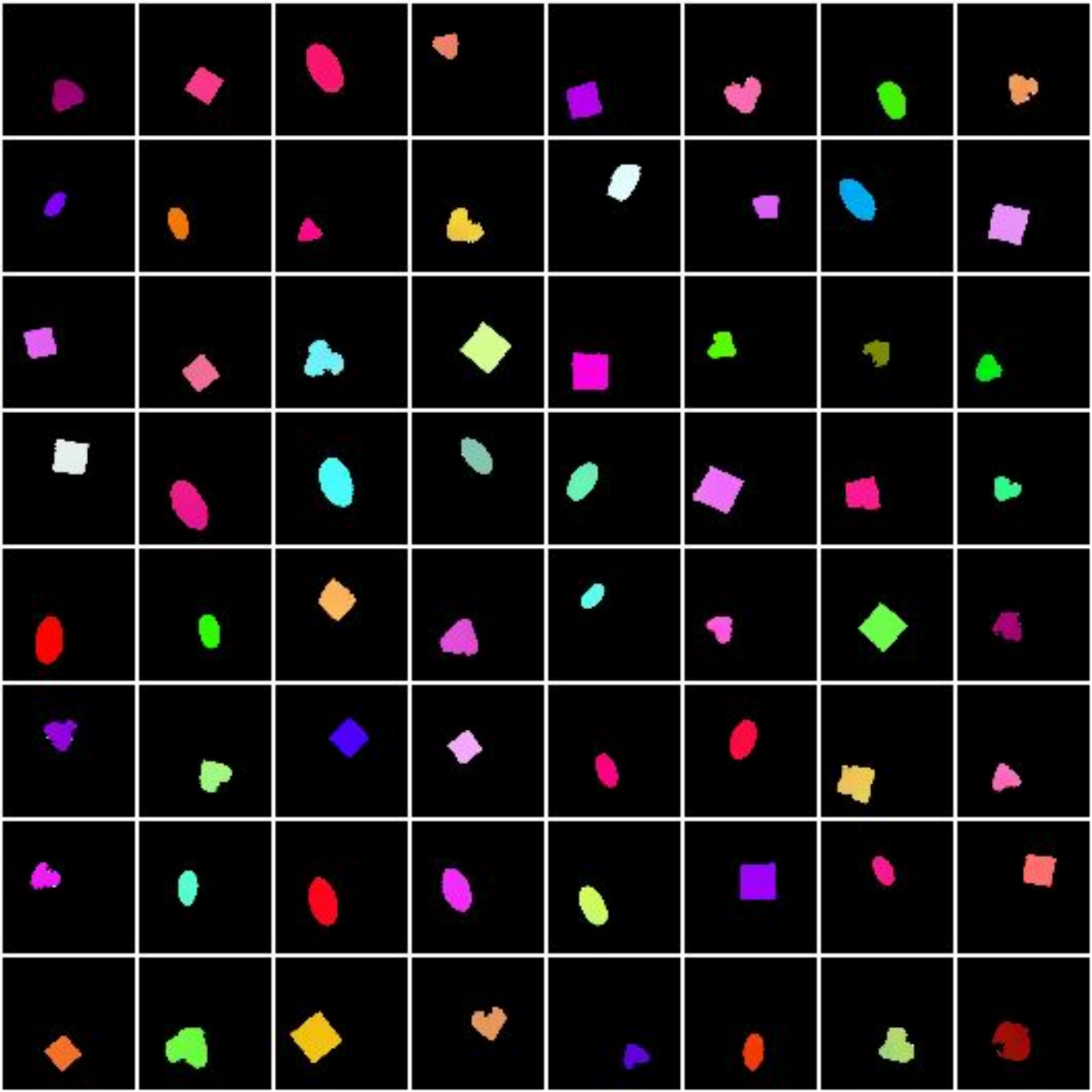}
        \caption[]
        {{\small Color-dSprites}}
    \end{subfigure}
    \vskip\baselineskip
    \begin{subfigure}[b]{0.38\textwidth}   
        \centering 
        \includegraphics[width=\textwidth]{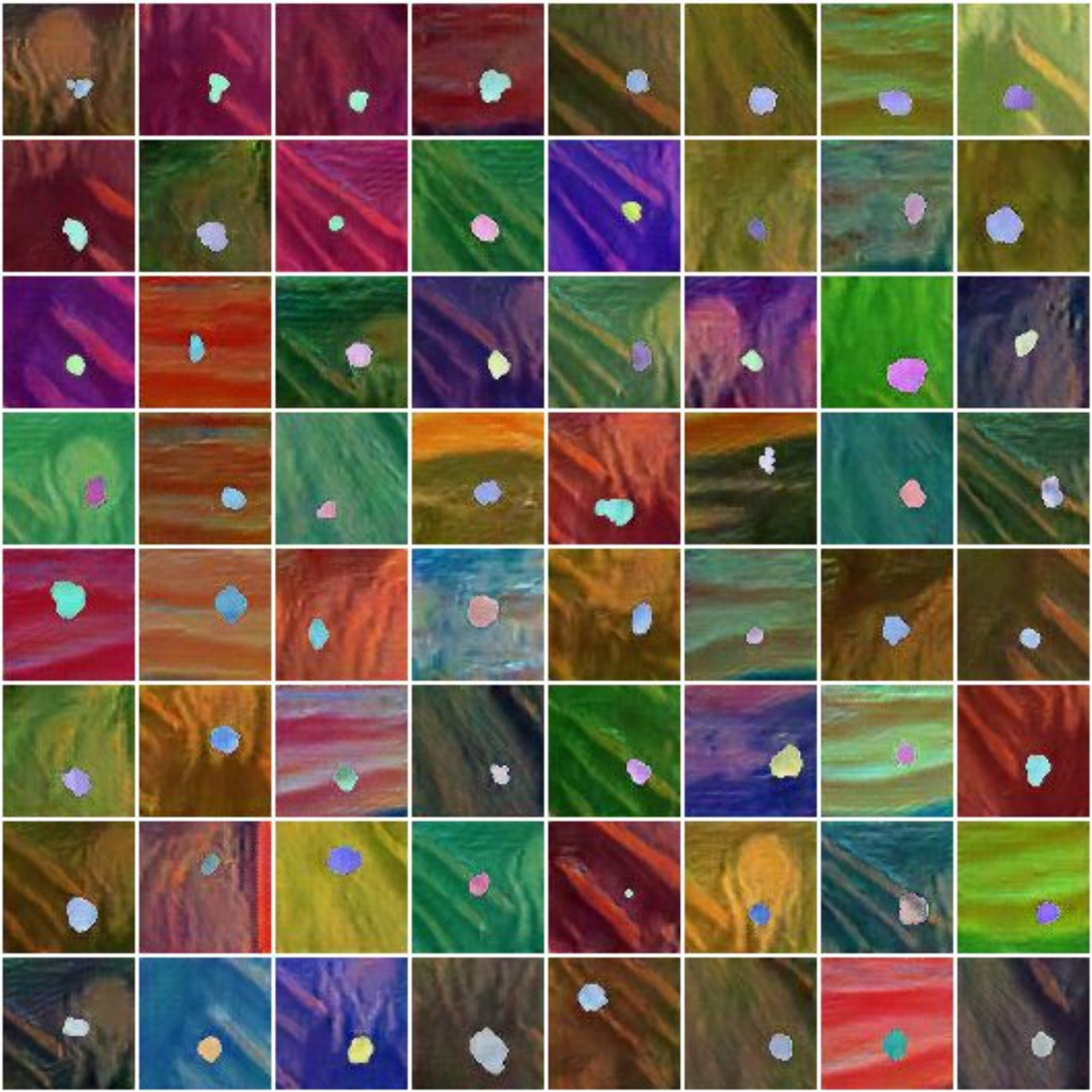}
        \caption[]
        {{\small Scream-dSprites}}
    \end{subfigure}
    \hspace{0.3in}
    \begin{subfigure}[b]{0.38\textwidth}   
        \centering 
        \includegraphics[width=\textwidth]{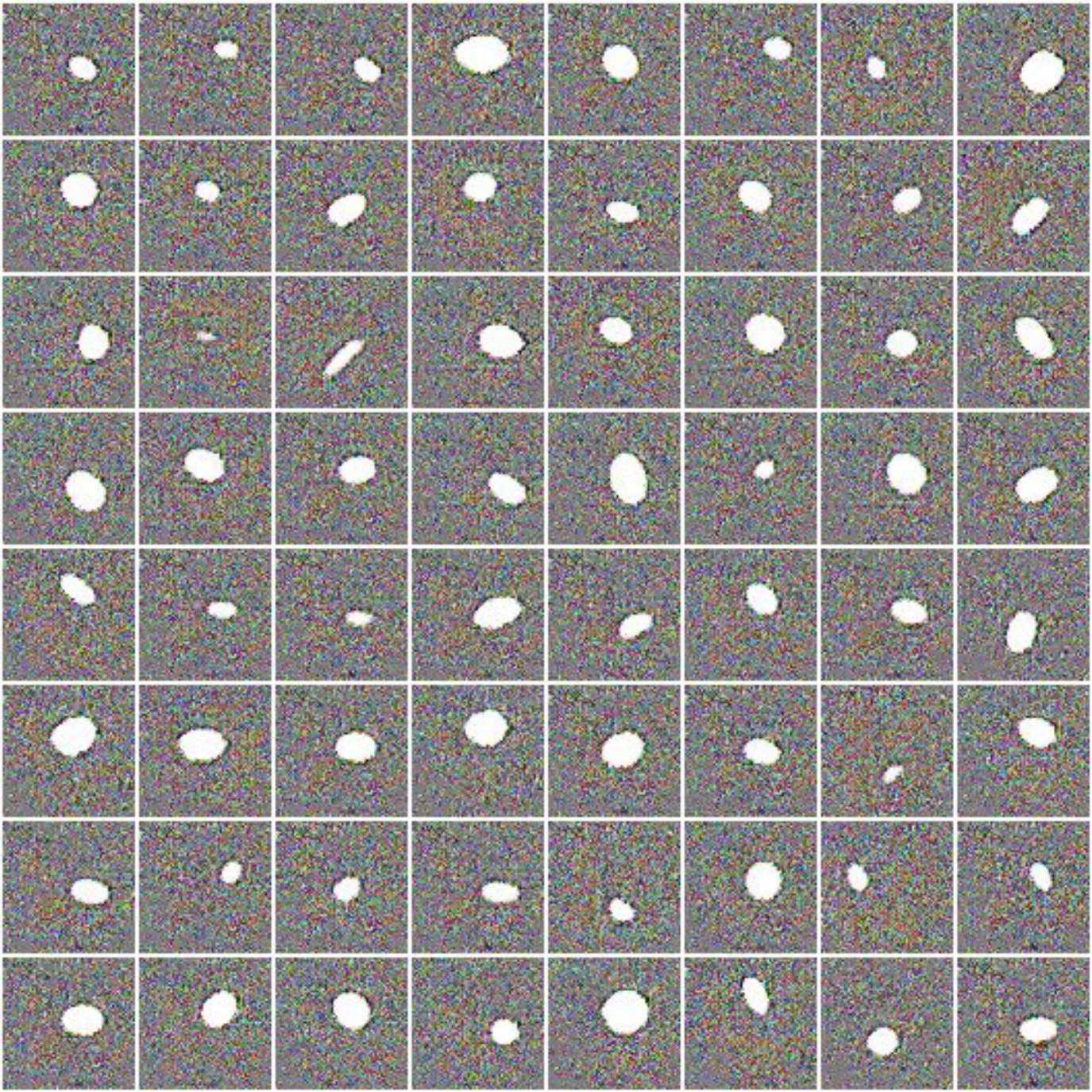}
        \caption[]
        {{\small Noisy-dSprites}}
    \end{subfigure}
    \caption[]
    {\small Random generated samples of ID-GAN on dSprites and its variants.} 
    \label{fig:idgan_synthetic_samples}
\end{figure*}

We present additional qualitative results on the synthetic dataset, which corresponds to Section \ref{sec:synthetic_experiments} in the main paper.
Figure~\ref{fig:idgan_synthetic_samples} presents the randomly generated images by the proposed ID-GAN.
We observe that the generated images are sharp and realistic, capturing complex patterns in the background (Screem-dSprites and Noise-dSprites datasets). 
We also observe that it generates convincing foreground patterns, such as color and shape of the objects, while covering diverse and comprehensive patterns in real objects.

\begin{figure*}[htbp!]
    \centering
    \begin{subfigure}[b]{\textwidth}
        \centering
        \includegraphics[width=0.8\textwidth]{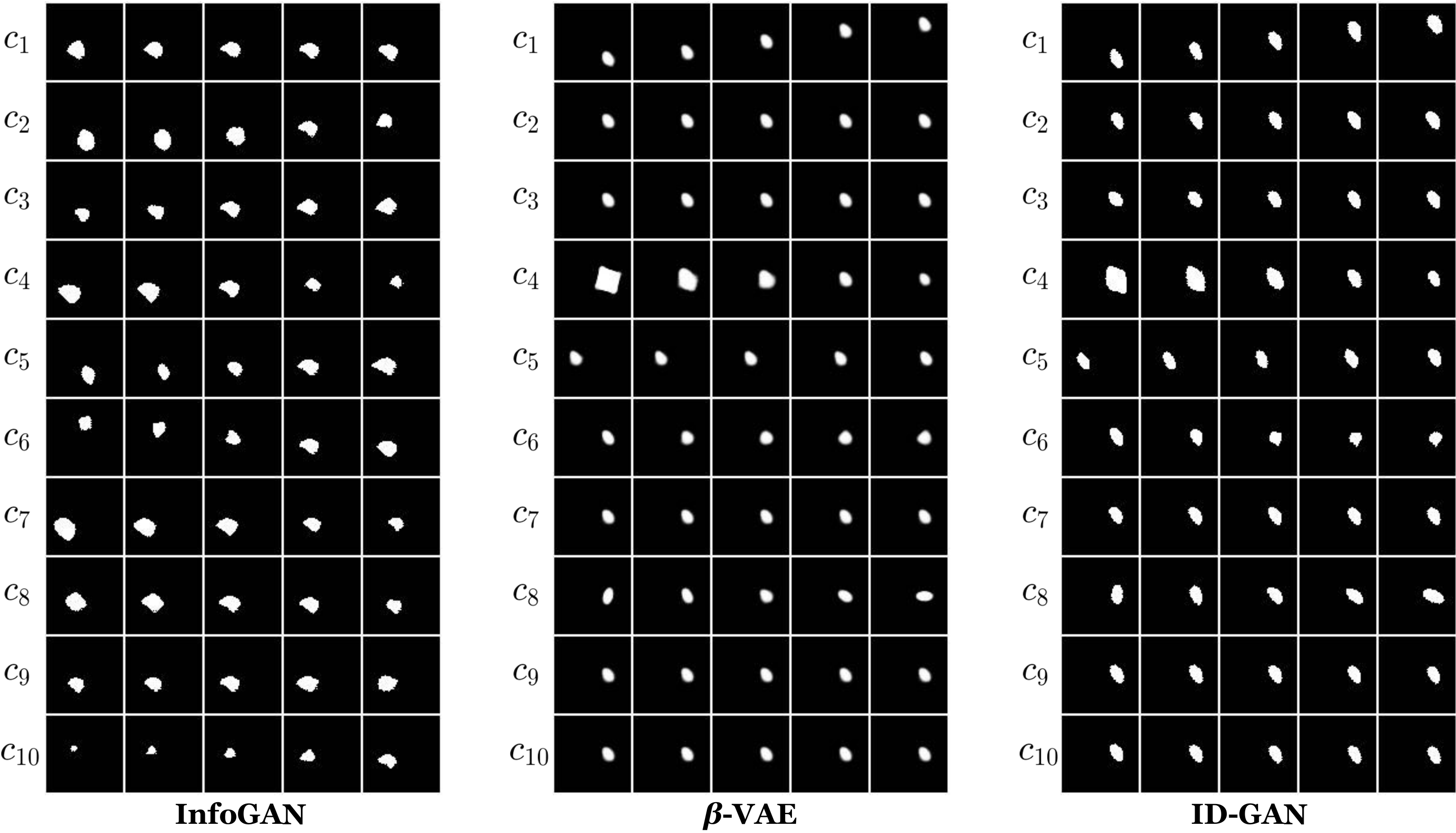}
        \caption[]
        {{\small dSprites}}
    \end{subfigure} \label{fig:full_latent_traversals_dsprites}
    \vskip\baselineskip\vspace{0.2cm}
    \begin{subfigure}[b]{\textwidth}
        \centering
        \includegraphics[width=0.8\textwidth]{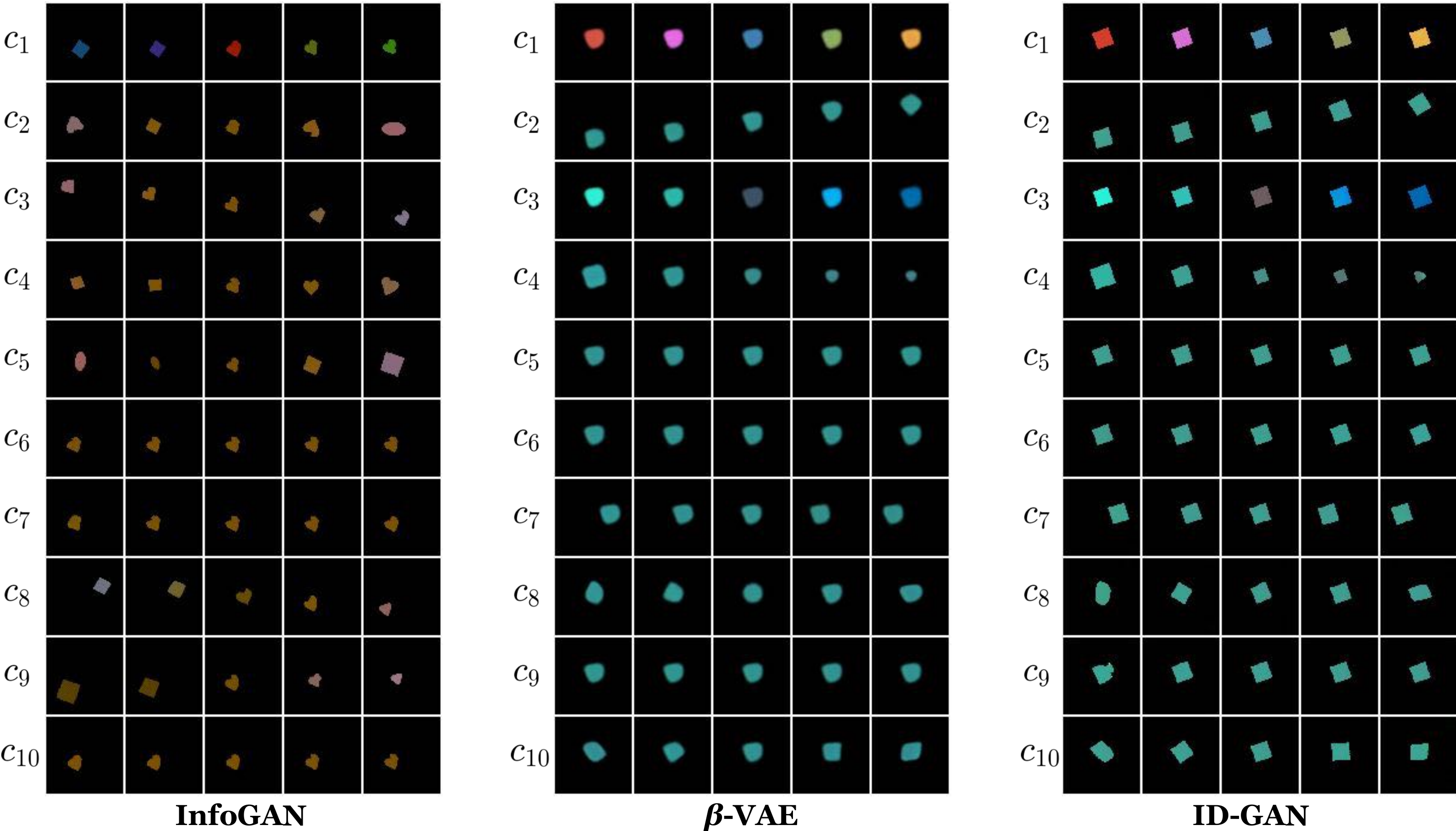}
        \caption[]
        {{\small Color-dSprites}}
    \end{subfigure} \label{fig:full_latent_traversals_cdsprites}
    \caption[ The average and standard deviation of critical parameters ]
    {\small Latent traversals of InfoGAN, $\beta$-VAE, and ID-GAN on (a) dSprites and (b) Color-dSprites datasets. Each $c_j$ $(j=1,\dots,10)$ represents a single dimension of $c$.}
    \label{fig:full_latent_traversals_1}
\end{figure*}

\begin{figure*}[htbp!]
    \centering
    \begin{subfigure}[b]{\textwidth}
        \centering
        \includegraphics[width=0.8\textwidth]{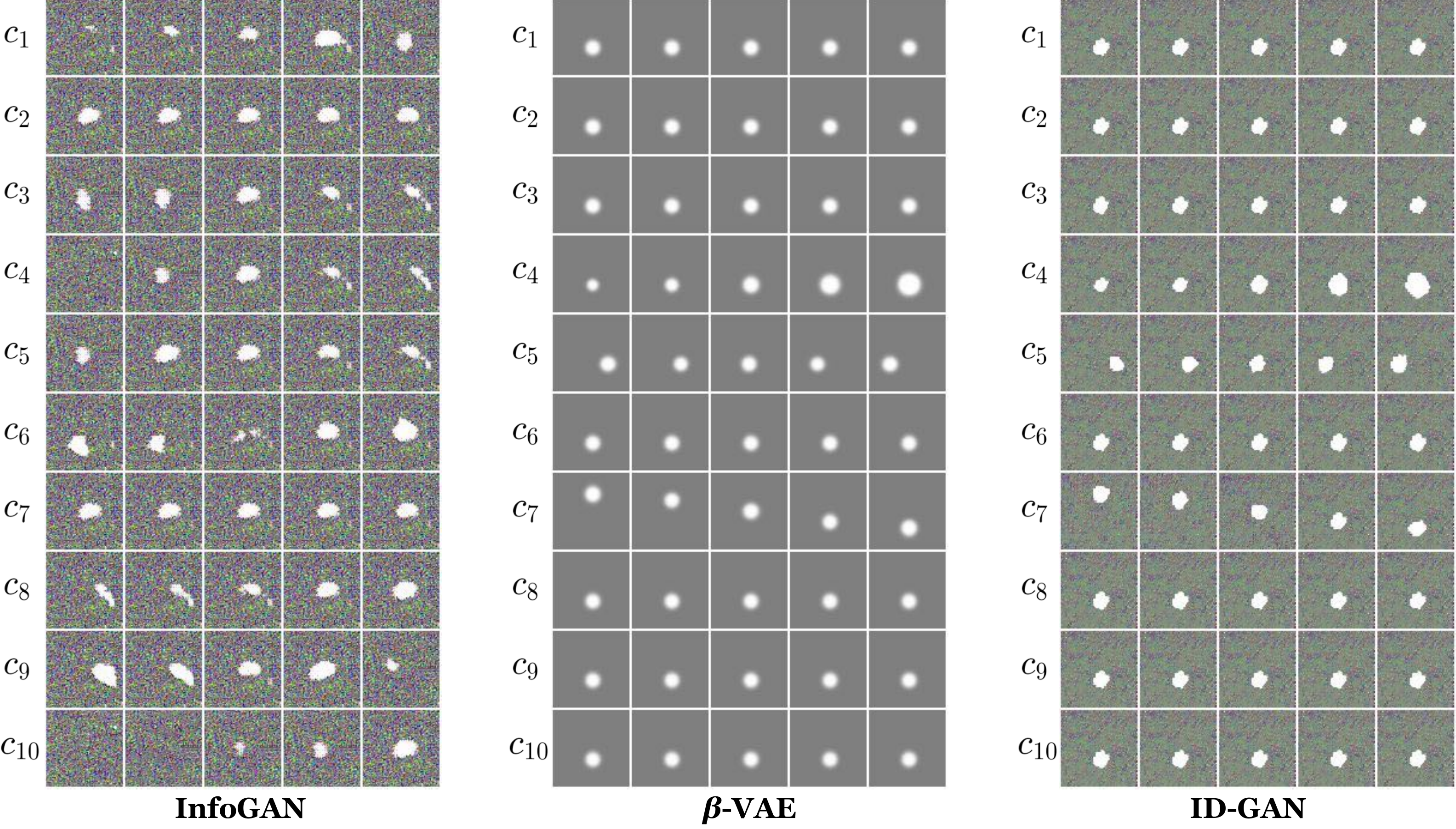}
        \caption[]
        {{\small Noisy-dSprites}}
    \end{subfigure} \label{fig:full_latent_traversals_ndsprites}
    \vskip\baselineskip\vspace{0.2cm}
    \begin{subfigure}[b]{\textwidth}
        \centering
        \includegraphics[width=0.8\textwidth]{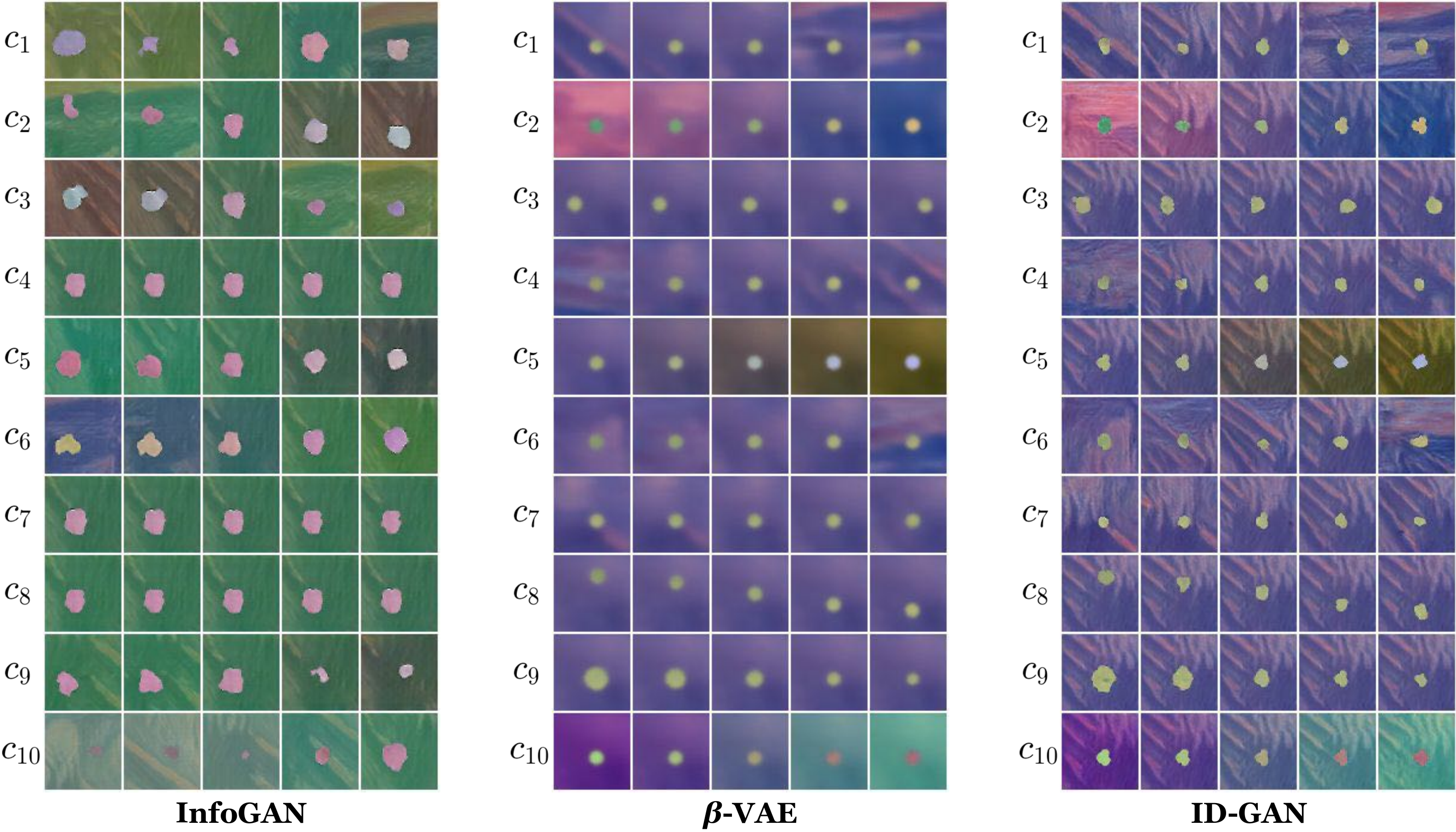}
        \caption[]
        {{\small Scream-dSprites}}
    \end{subfigure} \label{fig:full_latent_traversals_sdsprites}
    \caption[]
    {\small Latent traversals of InfoGAN, $\beta$-VAE, and ID-GAN on (a) Noisy-dSprites and (b) Scream-dSprites datasets. Each $c_j$ $(j=1,\dots,10)$ represents a single dimension of $c$.}
    \label{fig:full_latent_traversals_2}
\end{figure*}

Figure~\ref{fig:full_latent_traversals_1} and \ref{fig:full_latent_traversals_2} present additional qualitative comparison results with $\beta$-VAE and InfoGAN by manipulating the disentangled factors, which correspond to Figure \ref{fig:synthetic_qualitative_eval} in the main paper.  
We observe that $\beta$-VAE captures the meaningful disentangled factors, such as location and color of the object, but overlooks several complex but important patterns in the background (Screem-dSprites and Noise-dSprites datasets) as well as foreground (\eg~detailed shape and orientation).
On the other hand, InfoGAN generates more convincing images by employing a more expressive decoder, but learns much more entangled representations (\eg~changing location and color of the objects in Color-dSprites dataset). 
By combining the benefits of both approaches, ID-GAN successfully learns meaningful disentangled factors and generates realistic patterns.

\null\vfill
\vfill\clearpage

\null\vfill
\vfill\clearpage

\subsection{Additional Results on a Complex Dataset}
\begin{figure*}[htbp!]
    \centering
    \includegraphics[width=1\textwidth]{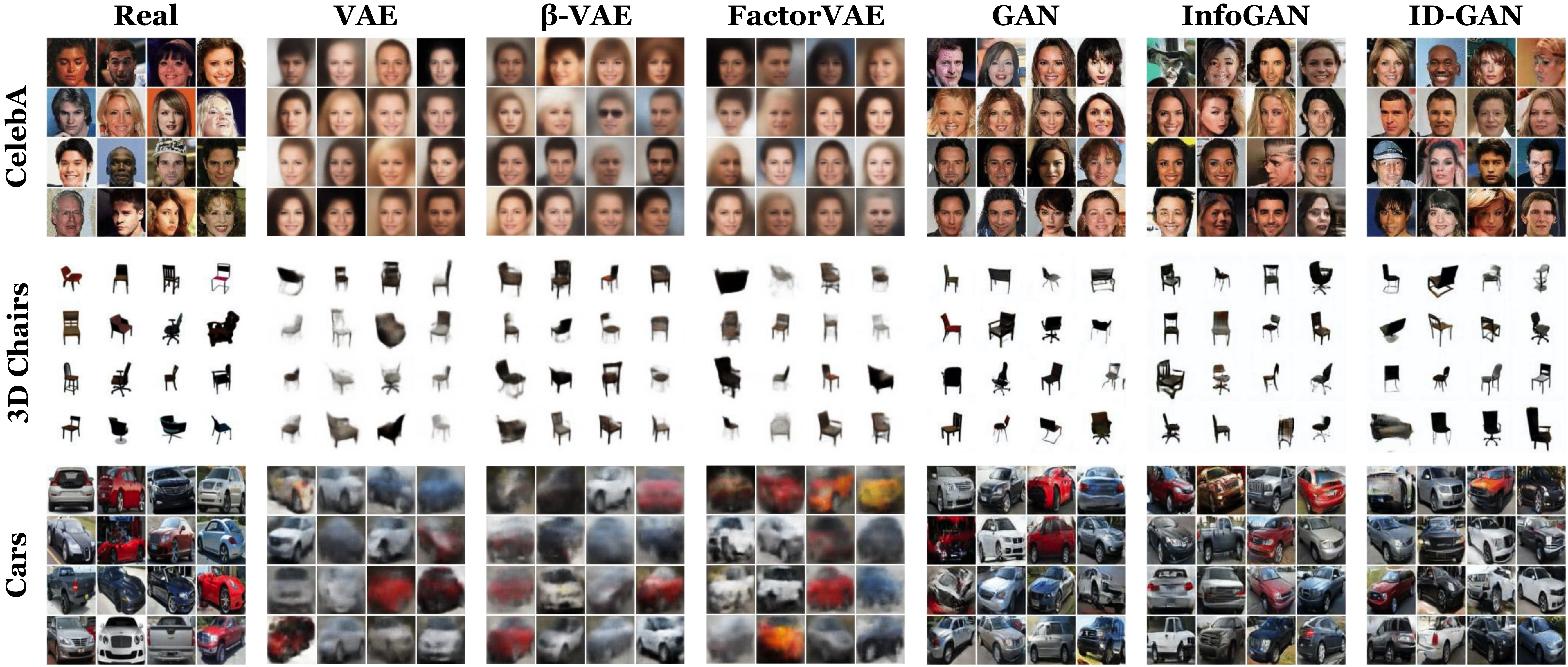}
    \caption[]
    {\small Random samples generated by VAE~\cite{misc/vae}, $\beta$-VAE~\cite{dsr/betavae}, FactorVAE~\cite{dsr/factorvae}, GAN~\cite{misc/ganbase}, InfoGAN~\cite{dsr/infogan}, and ID-GAN on CelebA, 3D Chairs, and Cars datasets~(64$\times$64).}
    \label{supp:fig:idgan_complexc_samples}
    \vspace{-0.1in}
\end{figure*}

\begin{figure*}[hbp!]
    \centering        
    \includegraphics[width=.735\textwidth]{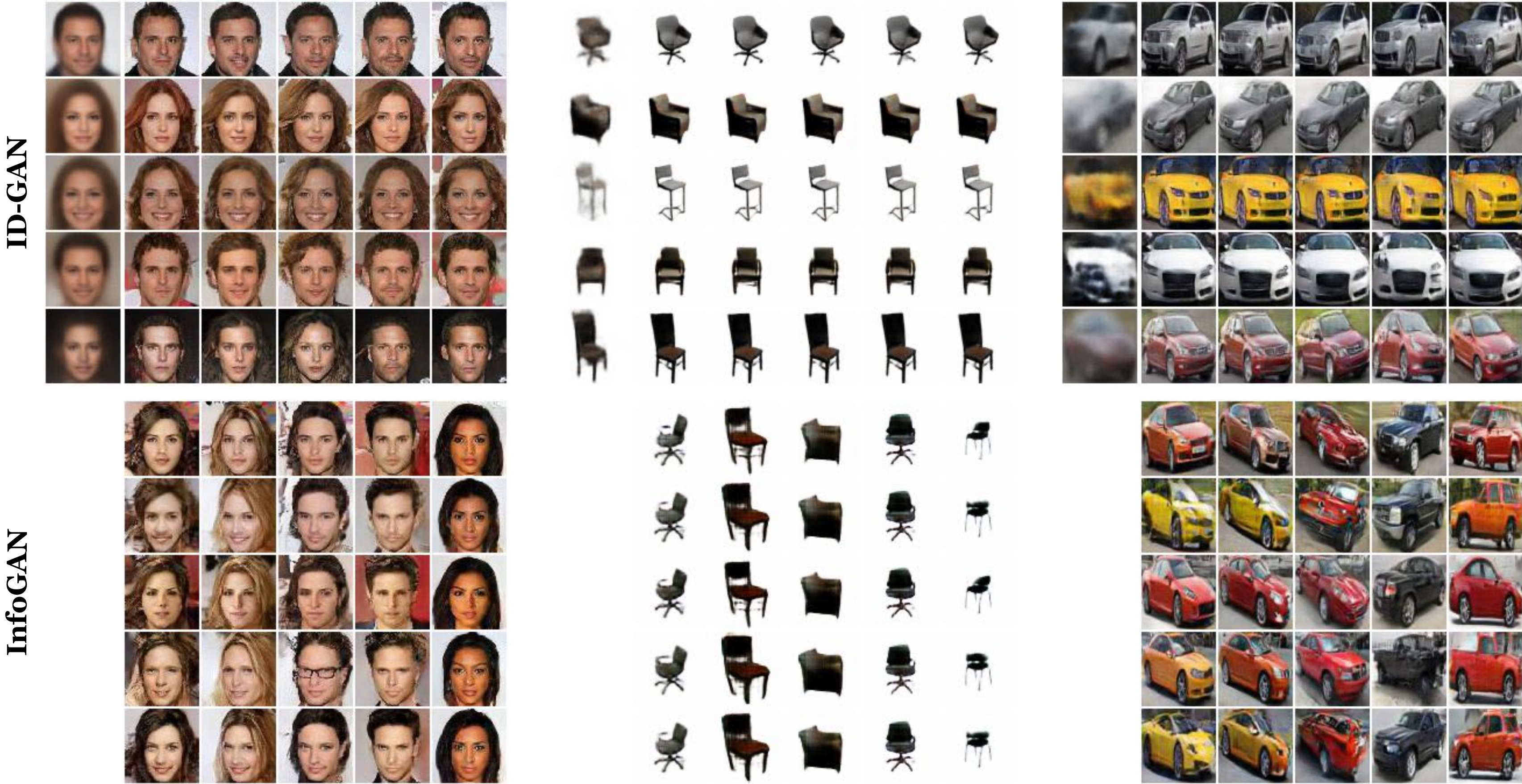}
    \caption[]
    {\small Analysis on the learned factors of variations in ID-GAN and InfoGAN. We sample 5 points for each $c$ and $s$ and visualize how they are used during the generative process of ID-GAN and InfoGAN. The samples in each row are generated from a single $c$ with 5 different $s$. Also, we show the generated samples of $\beta$-VAE from each $c$ at the first column of the panes of ID-GAN. As shown above, ID-GAN successfully learns to decode the global structures of data from $c$ as $\beta$-VAE, while utilizing $s$ as additional sources of variations for modeling local details. In InfoGAN, however, the most salient factors of variations are dominated by $s$ while $c$ acts as the nuisance.} %
    \label{supp:fig:infogan_z_dominate}
\end{figure*}

\paragraph{Qualitative results of Table 4.}
Here we compare the qualitative samples generated by each model in Table \ref{tab:complex_quantitative_eval}, \ie. VAE, $\beta$-VAE, FactorVAE, GAN, InfoGAN, and ID-GAN. The qualitative results are shown in Figure~\ref{supp:fig:idgan_complexc_samples}.
Although VAE-based methods learn to represent global structures or salient factors of data in all datasets, generated samples are often blurry and lack textural or local details.
On the other hand, GAN-based approaches (\ie~GAN, InfoGAN and ID-GAN) generate sharp and realistic samples thanks to the implicit density estimation and expressive generators.
However, as shown in Figure~\ref{supp:fig:infogan_z_dominate}, InfoGAN generally fails to capture meaningful disentangled factors into $c$ since it exploits the  nuisance variable $s$ to encode the most salient factors of variations.
On the other hand, ID-GAN successfully captures major disentangled factors into $c$ while encoding only local details into the nuisance variable $s$.

\clearpage

\begin{figure*}[htbp!]
    \centering
    \includegraphics[width=0.95\textwidth]{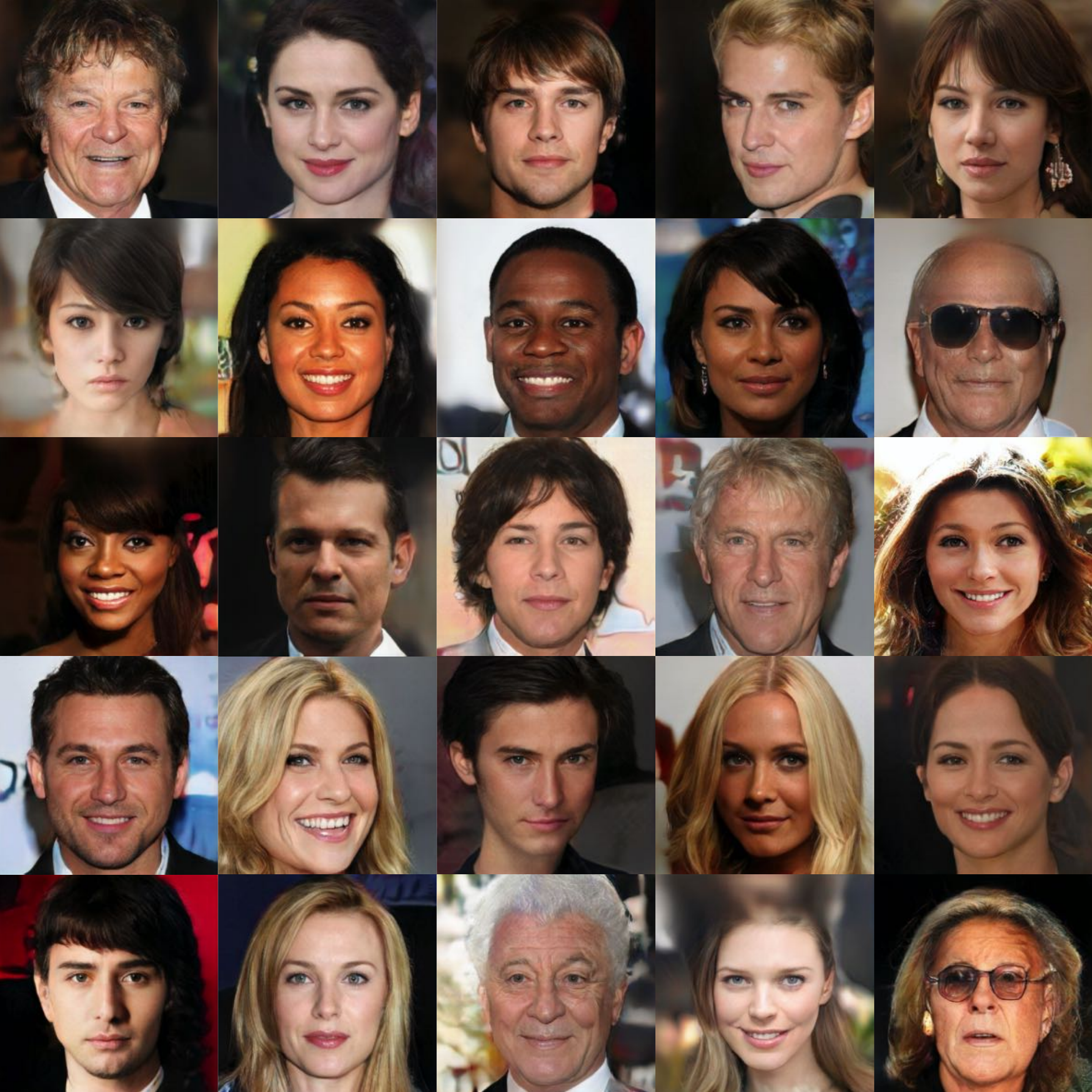}
    \caption{
    Random samples generated by ID-GAN on the CelebA-HQ dataset ($1024\times1024$). 
    ID-GAN is based on VGAN architecture~\cite{misc/ganbasevdb} and is trained to render learned disentangled representation $c$ of $\beta$-VAE trained on much smaller $64\times64$ image resolution.
    }
    \label{supp:fig:hq_random}
\end{figure*}

\paragraph{Additional results on high-resolution synthesis.}
Here we provide more qualitative results of ID-GAN on high-resolution image synthesis (CelebA 256$\times$256 and CelebA-HQ datasets).
We first present the results on the CelebA-HQ dataset composed of mega-pixel images (1024$\times$1024 pixels). 
Figure~\ref{supp:fig:hq_random} presents the randomly generated samples by ID-GAN.
We observe that ID-GAN produces sharp and plausible samples on high resolution images, showing on par generation performance with the state-of-the-art GAN baseline~\cite{misc/ganbasevdb} employed as a backbone network of ID-GAN.
We argue that this is due to the separate decoder and generator scheme adopted in ID-GAN, which is hardly achievable in the VAE-based approaches using a factorized Gaussian decoder for explicit maximization of data log-likelihood.

\begin{figure*}[htbp!]
    \centering
    \includegraphics[width=\textwidth]{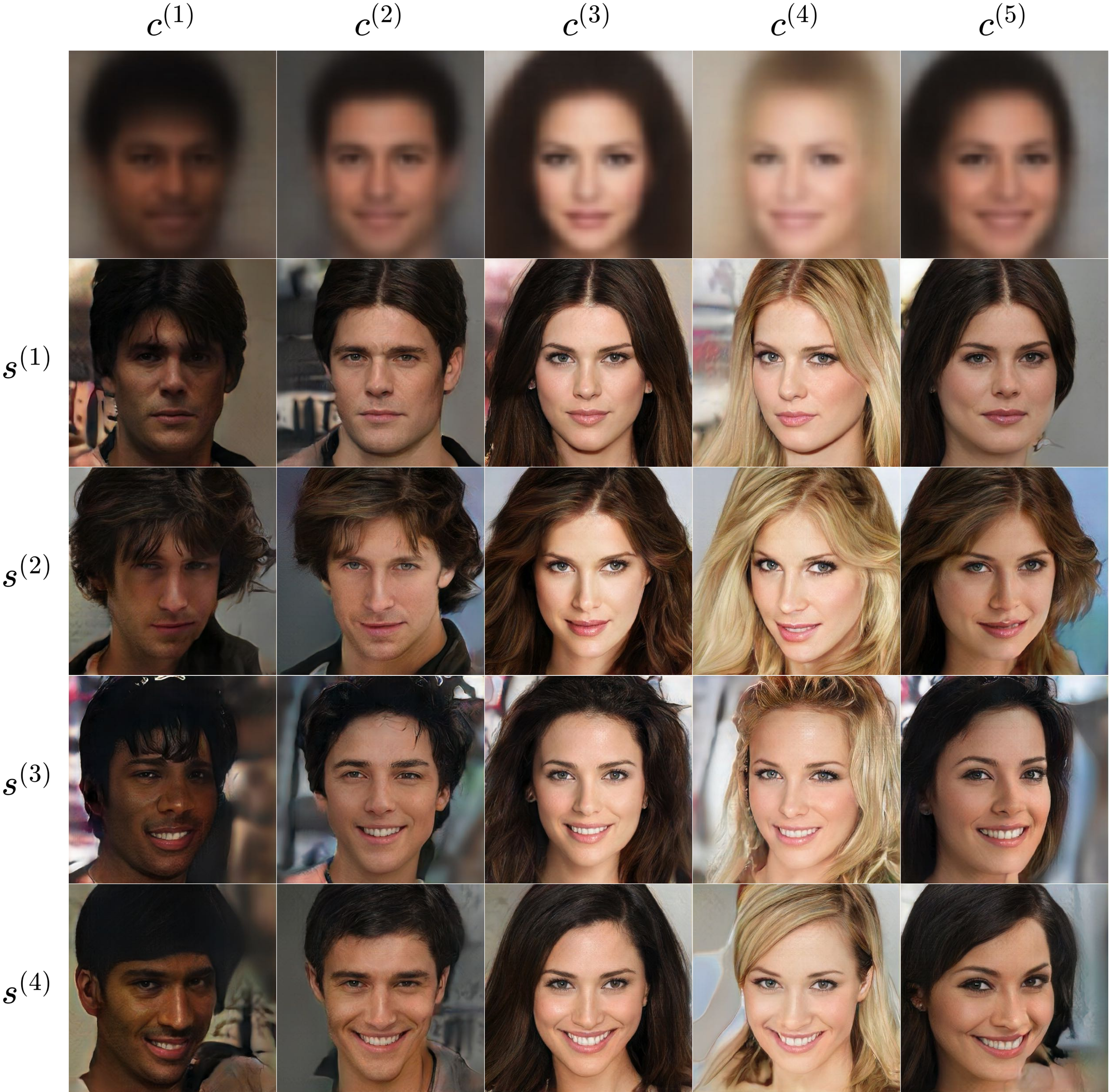}
    \caption{
    Similar visualizations as Figure~\ref{fig:complex_c_and_z_and_vae}, but on a more challenging CelebA-HQ ($1024\times1024$) dataset.
    We can clearly observe the different contributions of disentangled variable $c$ and nuisance variable $s$ to the generative process $G(s,c)$;
    disentangled variable $c$ captures the most salient factors of variations in the data (\eg azimuth and overall structure/color of face/hair are largely determined by $c$); 
    nuisance variable $s$ contributes to the remaining fine details (\eg identity, hair style, expression, background, \etc.).}
    \label{supp:fig:hq_factors}
\end{figure*}

Next, we analyze the learned factors of variations in ID-GAN by investigating the disentangled and nuisance variable $c$ and $s$, respectively.
Similarly to Figure {\color{red}8} in the main paper, we compare the samples generated by fixing one latent variable while varying another.
The results are summarized in Figure~\ref{supp:fig:hq_factors}.
Similar to Figure~\ref{fig:complex_c_and_z_and_vae}, we observe that the disentangled variable $c$ contributes to the most salient factors of variations (\eg azimuth, shape, or colour of face and hair, \etc.) while the nuisance variable $s$ contributes to the remaining fine details (\eg identity, hair style, expression, background, \etc.).
For instance, we observe that fixing the disentangled variable $c$ leads to consistent global face structure (\eg~black male facing slightly right (first column), blonde female facing slightly left (fourth column)), while fixing nuisance variable $s$ leads to consistent details (\eg~horizontally floating hair (third row), smiling expression (fourth and fifths rows)).
These results suggest that the generator in ID-GAN is well-aligned with the VAE decoder to render the same disentangled variable $c$ into similar observations, but with more expressive and realistic details by exploiting the nuisance variable $s$.

\begin{figure*}[htbp!]
    \centering
    \includegraphics[width=0.67\textwidth]{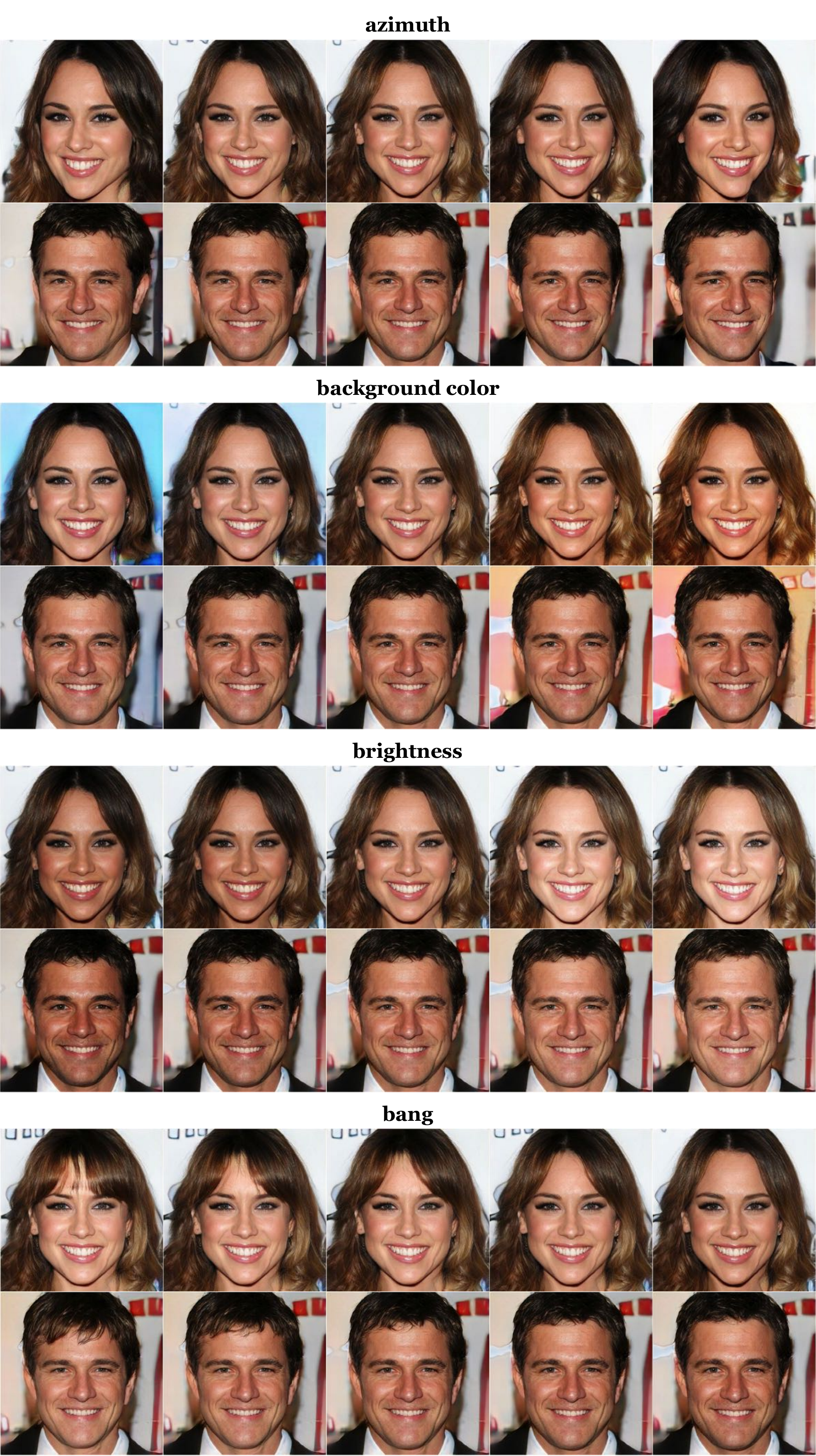}
    \caption{Latent traversal results of ID-GAN on CelebA-HQ ($1024\times1024$) dataset.}
    \label{supp:fig:hq_traverse}
\end{figure*}

\begin{figure*}[htbp!]
    \centering
    \includegraphics[width=0.8\textwidth]{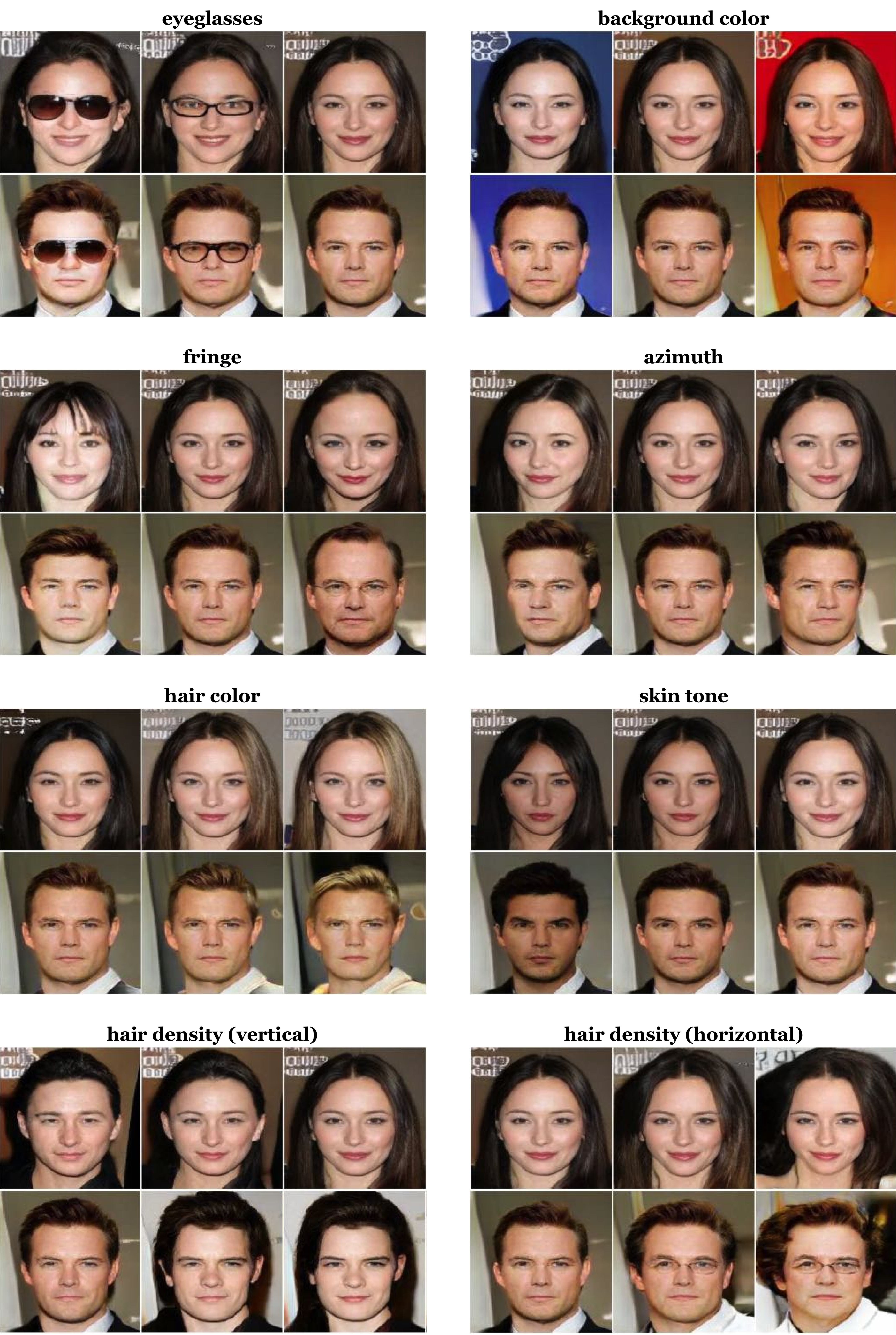}
    \caption{Latent traversals of ID-GAN on the CelebA dataset ($256\times256$).}
    \label{fig:sub_256_traverse}
\end{figure*}

Finally, to further visualize the learned disentangled factors in $c$, we present the latent traversal results in Figure~\ref{supp:fig:hq_traverse} as an extension to Figures~\ref{fig:hq_synthesis} and \ref{fig:complex_celebAHQ} in the main paper.
We also visualize the results on CelebA $256\times256$ images in Figure~\ref{fig:sub_256_traverse}, where we observe a similar behavior.
\wkk{

}
\vfill\clearpage

\null\vfill
\vfill\clearpage

\null\vfill
\vfill\clearpage

\null\vfill
\vfill\clearpage

\null\vfill

\null\vfill

\subsection{Sensitivity of Generation Performance (FID) on the Hyperparameter $\lambda$}
\label{supp:sensitivity}

\begin{figure}[ht!]
    \centering
    \includegraphics[width=0.6\textwidth]{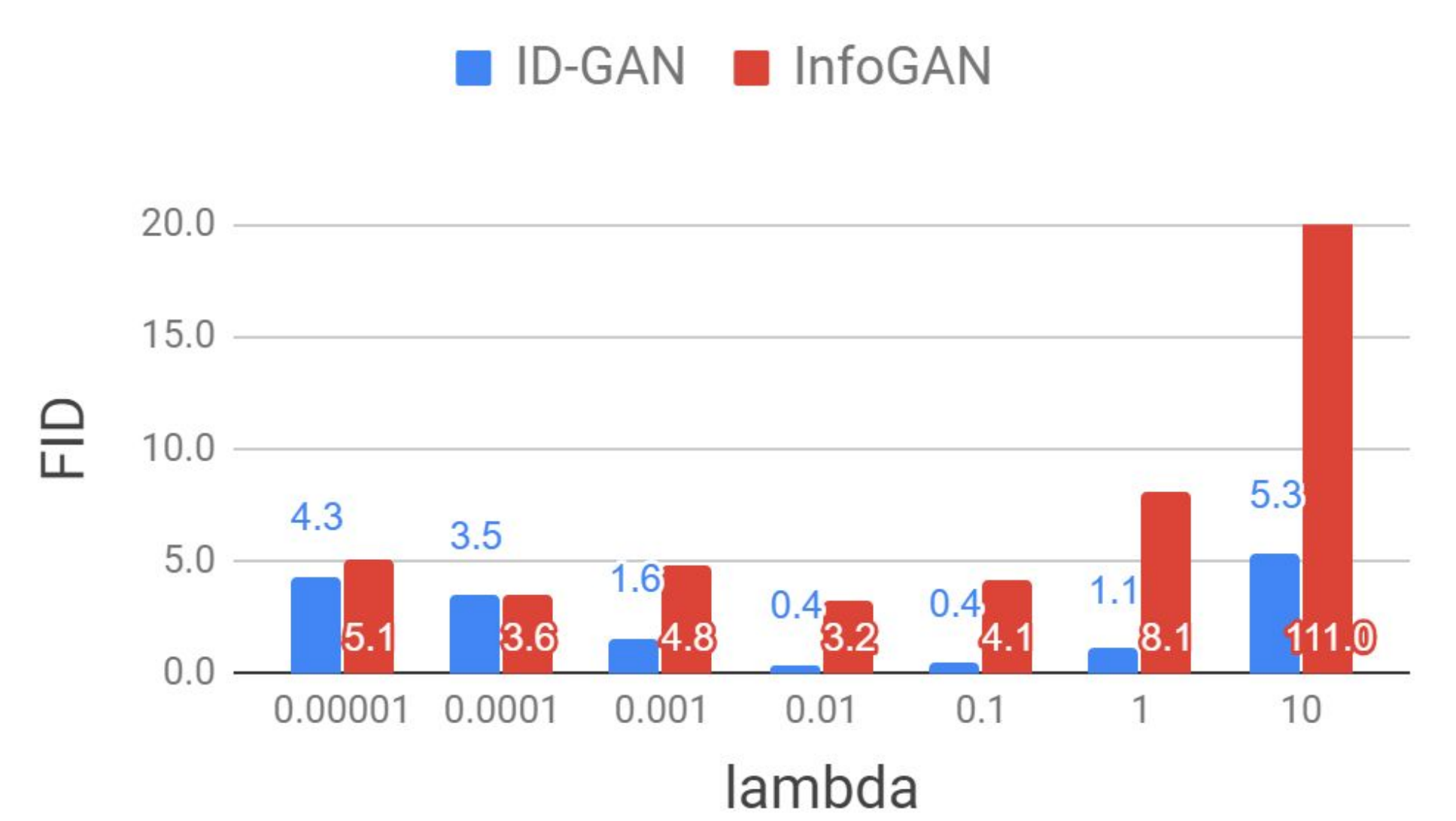}
    \caption[]{\small Sensitivity of the generation performance (FID) on $\lambda$.}
    \label{supp:fig:lamdba_sensitivity}
\end{figure}

To better understand the sensitivity of our model to its hyperparmeter ($\lambda$ in Eq.~(\ref{eqn:idgan})), we conduct an ablation study by measuring the generation performance (FID) of our models trained with various $\lambda$. 
Figure~\ref{supp:fig:lamdba_sensitivity} summarizes the results on the dSprite dataset.
First, we observe that the proposed ID-GAN performs well over a wide range of hyper-parameters ($\lambda\in[0.001, 1]$) while the performance of InfoGAN is affected much sensitively to the choice of $\lambda$.
Interestingly, increasing the $\lambda$ in our method also leads to the improved generation quality over a certain range of $\lambda$.
We suspect that it is because the information maximization in Eq.~(4) using the pre-trained encoder also behaves as the perceptual loss~\cite{perceptual_jj, perceptual_d, perceptual_lc}, regularizing the generator to match the true distribution in more meaningful feature space (\ie~disentangled representation). 
\vfill\clearpage

\section{Implementation Details}
\subsection{Evaluation Metrics}
\paragraph{FactorVAE Metric (FVM).} 
FVM~\cite{dsr/factorvae} measures the accuracy of a majority-vote classifier, where the encoder network to be evaluated is used for constructing the training data of this classifier. 
A single training data, or \textit{vote}, is generated as follows: 
we first extract encoder outputs from the entire samples of a synthetic dataset; 
estimate empirical variances of each latent dimension from the extracted outputs; 
sort out collapsed latent dimensions of variances smaller than 0.05; 
synthesize 100 samples with a single factor fixed and the other factors varying randomly; 
extract encoder outputs from synthesized samples;
compute variances of each latent dimension divided by the empirical variances computed beforehand;
then finally get a single vote which is a pair of the index of the fixed factor and the index of the latent dimension with the smallest normalized variance. 
We generate 800 votes to train the majority-vote classifier and report its train accuracy as the metric score.

\paragraph{Mutual Information Gap (MIG).}
MIG~\cite{dsr/betatcvae} is an information-theoretic approach to measure the disentanglement of representations. 
Specifically, assuming $K$ generative factors $v_k~(k=1,\dots,K)$ and $D$-dimensional latents $z_j~(j=1,\dots,D)$, it computes a normalized empirical mutual information $I(z_j;v_k)/H(v_k)$ to measure the information-theoretic preferences of $z_j$ towards each $v_k$, or vice versa. 
Then, it aggregates the differences, or \textit{gap}, between the top two preferences for each $v_k$ and averages them to compute MIG, \emph{i.e.} $\frac{1}{K}\sum_{k=1}^K\frac{1}{H(v_k)}(I(z_{j(k)};v_k)-\minl_{j \neq j(k)}{I(z_j;v_k)})$, where $I(z_{j(k)};v_k)=\argmax_j{I(z_j;v_k)}$. 
For implementation details, we directly follow the settings\footnote{https://github.com/google-research/disentanglement\_lib} in \cite{dsr/google}.

\paragraph{Fr\'echet Inception Distance (FID).}
We employ Fr\'echet Inception Distance (FID)~\cite{misc/fid} to evaluate the generation quality of each model considered in our experiments.
FID measures the Fréchet distance~\cite{misc/frechet} between two Gaussians, constructed by generated and real images, respectively, in the feature space of a pre-trained deep neural network. 
For each model, we compare 50,000 generated images and 50,000 real images to compute FID. 
For dSprites and its variants, we use a manually trained ConvNet trained to predict true generative factors of dSprites and its varaints. 
For the CelebA, 3D Chairs RGB, and Cars datasets, we use Inception V3~\cite{misc/inception} pre-trained on the ImageNet~\cite{misc/imagenet} dataset. 
We use the publicly available code\footnote{https://github.com/mseitzer/pytorch-fid} to compute FID.

\subsection{Dataset}
\begin{table}[htbp!]
\caption{Descriptions on datasets.}
\centering
\small
\begin{tabularx}{\textwidth}{lX}
\toprule
Name            & Description \\
\midrule
dSprites~\cite{dsr/dsprites}                    
& 737,280 binary 64x64 images of 2D sprites with 5 ground-truth factors, including shape~(3), scale~(6), orientation~(40), x-position~(32), and y-position~(32). \\
\midrule
Color-dSprites~\cite{dsr/annealedvae, dsr/google}
& The sprite is filled with a random color. We randomly sample intensities of each color channel from 8 discrete values, linearly spaced between [0, 1].\\
\midrule
Noisy-dSprites~\cite{dsr/google}                
& The background in each dSprites sample is filled with random uniform noise.\\
\midrule
Scream-dSprites~\cite{dsr/google}               
& The background of each dSprites sample is replaced with a randomly-cropped patch of \textit{The Scream} painting~\cite{} and the sprite is colored with the inverted color of the patch over the pixel regions of the sprite.\\
\midrule
CelebA, CelebA-HQ~\cite{dsr/celeba, dsr/celeba-hq}
& CelebA dataset contains 202,599 RGB images of celebrity faces, which is composed of 10,177 identities, 5 landmark locations, and 40 annotated attributes of human faces. We use the \textit{aligned\&cropped} version of the dataset with the image size of 64$\times$64 and 256$\times$256. CelebA-HQ is the subset of the \textit{in-the-wild} version of the CelebA dataset, which is composed of 30,000 RGB 1024$\times$1024 high-resolution images.\\
\midrule
3D Chairs~\cite{dsr/chairs}                     
& 86,366 RGB 64$\times$64 images of chair CAD models with 1,393 types, 31 azimuths, and 2 elevations.\\
\midrule
Cars~\cite{dsr/cars}                            
& 16,185 RGB images of 196 classes of cars. We crop and resize each image into the size of 64$\times$64 using the bounding-box annotations provided.\\
\bottomrule
\end{tabularx}
\end{table}

\subsection{Architecture}
\begin{table}[htbp!]
  \caption{Architectures of $\beta$-VAE and FactorVAE for all datasets. Note that Discriminator is needed only when training FactorVAE.}
  \vspace{.3cm}
  \centering
  \begin{tabular}{lll}
    \toprule
    \textbf{Encoder}                            & \textbf{Decoder}                          & \textbf{Discriminator} \\
    \midrule
    Input: 64 $\times$ 64 $\times$ \# channels  & Input: $\mathbb{R}^{10}$                  & FC 1000, leaky ReLU    \\
    4$\times$4 conv 32, ReLU, stride 2          & FC 256, ReLU                              & FC 1000, leaky ReLU    \\
    4$\times$4 conv 32, ReLU, stride 2          & FC 4$\times$4$\times$64, ReLU             & FC 1000, leaky ReLU    \\
    4$\times$4 conv 64, ReLU, stride 2          & 4$\times$4 upconv 64, ReLU, stride 2      & FC 1000, leaky ReLU    \\
    4$\times$4 conv 64, ReLU, stride 2          & 4$\times$4 upconv 32, ReLU, stride 2      & FC 1000, leaky ReLU    \\
    FC 256, FC 2$\times$10                      & 4$\times$4 upconv 32, ReLU, stride 2      & FC 1000, leaky ReLU    \\
                                                & 4$\times$4 upconv \# channels, stride 2   & FC 2                   \\
    \bottomrule
    \label{table:dvae_arch}
  \end{tabular}
\end{table}

\begin{table}[htbp!]
  \caption{Architectures of Generator and Discriminator networks for ID-GAN and InfoGAN on dSprites, Color-dSprites, Noisy-dSprites, and Scream-dSprites datasets. The encoder and the decoder networks are specified in Table~\ref{table:dvae_arch}.}
  \vspace{.4cm}
  \centering
  \begin{tabular}{ll}
    \toprule
    \textbf{Generator}                          & \textbf{Discriminator} \\
    \midrule
    Input: $\mathbb{R}^{10}$                  & Input: 64 $\times$ 64 $\times$ \# channels    \\
    FC 256, ReLU                              & 4$\times$4 conv 32, ReLU, stride 2            \\
    FC 4$\times$4$\times$64, ReLU             & 4$\times$4 conv 32, ReLU, stride 2            \\
    4$\times$4 upconv 64, ReLU, stride 2      & 4$\times$4 conv 64, ReLU, stride 2            \\
    4$\times$4 upconv 32, ReLU, stride 2      & 4$\times$4 conv 64, ReLU, stride 2            \\
    4$\times$4 upconv 32, ReLU, stride 2      & FC 256, FC 1                                  \\
    4$\times$4 upconv \# channels, stride 2   &                                               \\
    \bottomrule
  \end{tabular}
\end{table}

\begin{table}[htbp!]
  \caption{Architectures of Generator and Discriminator networks for ID-GAN and InfoGAN on CelebA, 3D Chairs, and Cars ($64\times64$) datasets. 
  We directly follow the architecture proposed in~\cite{misc/ganbase}.
  The encoder and the decoder networks are specified in Table~\ref{table:dvae_arch}.}
  \vspace{.3cm}
  \centering
  \begin{tabular}{ll}
    \toprule
    \textbf{Generator}                        & \textbf{Discriminator} \\
    \midrule
    Input: $\mathbb{R}^{20+256}$              & Input: 64 $\times$ 64 $\times$ 3 \\
    FC 4$\times$4$\times$512                  & 3$\times$3 conv 64, stride 1     \\
    ResBlock 512, NN Upsampling               & ResBlock 64, AVG Pooling         \\
    ResBlock 256, NN Upsampling               & ResBlock 128, AVG Pooling        \\
    ResBlock 128, NN Upsampling               & ResBlock 256, AVG Pooling        \\
    ResBlock  64, NN Upsampling               & ResBlock 512, AVG Pooling        \\
    ResBlock  64, 4$\times$4 conv 3, stride 1 & FC 1                             \\
    \bottomrule
  \end{tabular}
\end{table}

\begin{table}[htbp!]
  \caption{Architectures of Generator and Discriminator networks for ID-GAN (w/o distill), cGAN, and ID-GAN on the CelebA~(128$\times$128) dataset. 
  We directly follow the architecture proposed in~\cite{misc/ganbase}.
  The encoder and the decoder networks are specified in Table~\ref{table:dvae_arch}.}
  \vspace{.3cm}
  \centering
  \begin{tabular}{ll}
    \toprule
    \textbf{Generator}                        & \textbf{Discriminator} \\
    \midrule
    Input: $\mathbb{R}^{20+256}$              & Input: 128 $\times$ 128 $\times$ 3 \\
    FC 4$\times$4$\times$512                  & 3$\times$3 conv 64, stride 1       \\
    ResBlock 512, NN Upsampling               & ResBlock 64, AVG Pooling           \\
    ResBlock 512, NN Upsampling               & ResBlock 128, AVG Pooling          \\
    ResBlock 512, NN Upsampling               & ResBlock 256, AVG Pooling          \\
    ResBlock 256, NN Upsampling               & ResBlock 512, AVG Pooling          \\
    ResBlock 128, NN Upsampling               & ResBlock 512, AVG Pooling          \\
    ResBlock 128, 4$\times$4 conv 3, stride 1 & FC 1                               \\
    \bottomrule
  \end{tabular}
\end{table}

\begin{table}[htbp!]
  \caption{Architectures of Generator and Discriminator networks for ID-GAN on the CelebA~(256$\times$256) dataset. 
  We directly follow the architecture proposed in~\cite{misc/ganbase}.
  The encoder and the decoder networks are specified in Table~\ref{table:dvae_arch}.}
  \vspace{.3cm}
  \centering
  \begin{tabular}{ll}
    \toprule
    \textbf{Generator}                        & \textbf{Discriminator} \\
    \midrule
    Input: $\mathbb{R}^{20+256}$              & Input: 256 $\times$ 256 $\times$ 3 \\
    FC 4$\times$4$\times$512                  & 3$\times$3 conv 64, stride 1       \\
    ResBlock 512, NN Upsampling               & ResBlock 64, AVG Pooling           \\
    ResBlock 512, NN Upsampling               & ResBlock 128, AVG Pooling          \\
    ResBlock 512, NN Upsampling               & ResBlock 256, AVG Pooling          \\
    ResBlock 256, NN Upsampling               & ResBlock 512, AVG Pooling          \\
    ResBlock 128, NN Upsampling               & ResBlock 512, AVG Pooling          \\
    ResBlock  64, NN Upsampling               & ResBlock 512, AVG Pooling          \\
    ResBlock  64, 4$\times$4 conv 3, stride 1 & FC 1                               \\
    \bottomrule
  \end{tabular}
\end{table}

\begin{table}[htbp!]
  \caption{Architectures of Generator and Discriminator networks for ID-GAN on the CelebA-HQ~(1024$\times$1024) dataset. 
  We directly follow the architecture proposed in~\cite{misc/ganbasevdb}. 
  The encoder and the decoder networks are specified in Table~\ref{table:dvae_arch}. }
  \vspace{.3cm}
  \centering
  \begin{tabular}{ll}
    \toprule
    \textbf{Generator}                        & \textbf{Discriminator} \\
    \midrule
    Input: $\mathbb{R}^{20+256}$              & Input: 1024 $\times$ 1024 $\times$ 3  \\
    FC 4$\times$4$\times$512                  & ResBlock 16, AVG Pooling        \\
    ResBlock 512, NN Upsampling               & ResBlock 32, AVG Pooling        \\
    ResBlock 512, NN Upsampling               & ResBlock 64, AVG Pooling        \\
    ResBlock 512, NN Upsampling               & ResBlock 128, AVG Pooling       \\
    ResBlock 512, NN Upsampling               & ResBlock 256, AVG Pooling       \\
    ResBlock 256, NN Upsampling               & ResBlock 512, AVG Pooling       \\
    ResBlock  128, NN Upsampling              & ResBlock 512, AVG Pooling       \\
    ResBlock  64, NN Upsampling               & ResBlock 512, AVG Pooling       \\
    ResBlock  32, NN Upsampling               & 1$\times$1 conv 2$\times$512, Sampling 512    \\
    ResBlock  16,4$\times$4 conv 3, stride 1  & FC 1                    \\
    \bottomrule
  \end{tabular}
\end{table}

\end{document}